\documentclass[11pt]{article}

\usepackage{tgtermes}

\usepackage[top=1.0in,bottom=1.0in,left=1.0in,right=1.0in]{geometry}

\usepackage{color}
\usepackage[table]{xcolor}
\usepackage{booktabs}

\usepackage{newtxtext}
\usepackage{newtxmath}
\usepackage{bm}

\usepackage{algorithm}
\usepackage{algpseudocode}
\usepackage{subcaption}
\usepackage{graphicx}
\usepackage{threeparttable}

\usepackage{natbib}
 \bibpunct[, ]{(}{)}{,}{a}{}{,}%

\title{
\raggedright
\Large \textbf{Online Kernel Dynamic Mode Decomposition for Streaming Time Series Forecasting with Adaptive Windowing} \\

\vspace{0.25in}
\large \textbf{Christopher Salazar}, Department of Industrial \& Systems Engineering, University of Washington, Seattle WA 98195, USA \\

\vspace{0.25in}
\large \textbf{Krithika Manohar}, Department of Mechanical Engineering, University of Washington, Seattle WA 98195, USA \\

\vspace{0.25in}
\large \textbf{Ashis G. Banerjee}, Department of Industrial \& Systems Engineering and Mechanical Engineering, University of Washington, Seattle WA 98195, USA \\
}

\date{}

\begin{document}

\maketitle

\section*{Abstract}
\label{sec:Abstract}

Real-time forecasting from streaming data poses critical challenges: handling non-stationary dynamics, operating under strict computational limits, and adapting rapidly without catastrophic forgetting. However, many existing approaches face trade-offs between accuracy, adaptability, and efficiency, particularly when deployed in constrained computing environments. We introduce WORK-DMD (Windowed Online Random Kernel Dynamic Mode Decomposition), a method that combines Random Fourier Features with online Dynamic Mode Decomposition to capture nonlinear dynamics through explicit feature mapping, while preserving fixed computational cost and competitive predictive accuracy across evolving data. WORK-DMD employs Sherman–Morrison updates within rolling windows, enabling continuous adaptation to evolving dynamics from only current data, eliminating the need for lengthy training or large storage requirements for historical data. Experiments on benchmark datasets across several domains show that WORK-DMD achieves higher accuracy than several state-of-the-art online forecasting methods, while requiring only a single pass through the data and demonstrating particularly strong performance in short-term forecasting. Our results show that combining kernel evaluations with adaptive matrix updates achieves strong predictive performance with minimal data requirements. This sample efficiency offers a practical alternative to deep learning for streaming forecasting applications.

\section{Introduction}
\label{sec:Introduction}

Real-time decision making in modern applications increasingly depends on accurate time series forecasting from continuously streaming data. From financial trading systems requiring millisecond response times to smart grid management adapting to fluctuating energy demand, the ability to generate reliable predictions from evolving data streams has become critical~\citep{lim2021time}. However, streaming data presents fundamental challenges that traditional forecasting methods struggle to address: non-stationarity, where the underlying patterns shift over time; stringent computational constraints that prohibit extensive model retraining; and catastrophic forgetting where models lose knowledge of past patterns when adapting to new data~\citep{kirkpatrick2017overcoming}. These challenges demand forecasting approaches that can adapt quickly to changing dynamics while maintaining computational efficiency and preserving predictive accuracy.

\subsection{Modern Approaches and Their Limitations}

Deep learning approaches, including recurrent networks~\citep{salinas2020deepar}, transformer-based models~\citep{Zhou2021Informer}, and recent foundation models~\citep{ansari2024chronos, woo2024moirai}, have demonstrated impressive forecasting capabilities in various domains. State-of-the-art models achieve remarkable zero-shot performance by leveraging massive pre-training on billions of time series observations~\citep{das2023decoder}. However, these methods incur substantial computational overhead, requiring extensive training over multiple epochs and large memory footprints that make real-time deployment challenging. More critically, they suffer from catastrophic forgetting in online settings. When fine-tuned on streaming data, these models lose previously learned patterns~\citep{buzzega2020dark}. Addressing this requires careful regularization strategies or replay mechanisms, which further increase computational demands.

Beyond computational constraints, deep learning models present fundamental interpretability challenges that complicate their deployment in critical applications. These black-box models offer limited insight into their decision-making processes~\citep{guidotti2018survey}, making it difficult for practitioners to understand when and why they succeed or fail on particular datasets. Recent work has sought to address this opacity through distance correlation-based approaches that characterize information flow through neural network layers~\citep{salazar2024distance}, revealing how RNNs gradually lose temporal information across activation layers and struggle with certain time series characteristics. Such interpretability challenges, combined with their computational demands, underscore the need for more transparent and efficient alternatives for streaming forecasting scenarios.

Foundation models represent the current frontier in time series forecasting, with models such as TimesFM~\citep{das2023decoder} and Moirai~\citep{woo2024moirai}, trained on 307 billion and 27 billion observations, respectively, demonstrating impressive zero-shot capabilities across diverse domains. Although these models excel in offline evaluation scenarios, their deployment in real-time streaming applications remains challenging due to their computational requirements and the difficulty of efficiently adapting their learned representations to evolving data distributions. 

Modern online learning methods have emerged to address some of these limitations. Approaches such as FSNet~\citep{pham2022learning} leverage frequency domain representations for multi-scale temporal modeling, while methods such as OneNet~\citep{zhang2023onenet} propose unified frameworks for diverse forecasting scenarios. Continual learning techniques such as DER++~\citep{buzzega2020dark} attempt to mitigate catastrophic forgetting through experience replay and regularization. However, these methods typically still require substantial training data, limiting their applicability in resource-constrained streaming environments.

\subsection{Dynamic Mode Decomposition Approaches}

Standard Dynamic Mode Decomposition (DMD) methods have proven highly effective for analyzing dynamical systems by decomposing complex temporal behaviors into interpretable modes and eigenvalues~\citep{schmid2010dmd}. DMD assumes that the dynamics of the underlying system remains stationary, enabling the identification of fixed linear operators that govern the temporal evolution. However, this stationarity assumption becomes a critical limitation in streaming scenarios where system dynamics evolve continuously. When applied to non-stationary time series, traditional DMD fails to capture regime changes and may produce misleading eigenvalue decompositions that no longer reflect current system behavior~\citep{proctor2016dynamic}.

To address these limitations in streaming contexts, several online DMD variants have been developed. Streaming DMD ~\citep{hemati2014dynamic} introduced a method for large datasets, enabling incremental updates as new data become available. Similarly, Liew et al.~\citep{liew2022streaming} demonstrated streaming DMD applications for short-term wind farm forecasting, showing practical benefits in domain-specific applications. However, these approaches remain fundamentally limited by their linear nature and may struggle to effectively model the complex nonlinear dynamics present in broader real-world time series. In addition, these methods encounter computational challenges that require incremental matrix expansion and memory costs that increase continuously as more data arrives.

Zhang et al.~\citep{zhang2019online} proposed a more sophisticated online DMD approach that addresses some computational efficiency concerns through Sherman-Morrison matrix updates, demonstrating effectiveness for time-varying dynamical systems. While this method provides a solid foundation for online DMD updates, it has not been extended to kernel methods and lacks evaluation on general time series forecasting tasks.

Kernel-based extensions of DMD attempt to capture nonlinear dynamics by implicitly mapping data to high-dimensional feature spaces~\citep{williams2015data}. Although theoretically appealing, these methods face computational scalability issues in streaming contexts. Scalability requirements of kernel methods are fundamentally limited by intensive computations, where the covariance matrix needs to be inverted to compute posterior distributions, with computational complexity that grows with the number of points in the training data ~\citep{karsmakers2017scalable}. This quadratic scaling in memory and computation creates particular challenges for streaming applications, where maintaining and updating kernel matrices becomes increasingly expensive as new data continuously arrive.

\subsection{Our Approach and Contributions}

To address these limitations, we propose WORK-DMD (Windowed Online Random Kernel Dynamic Mode Decomposition), a novel approach that combines the mathematical rigor of DMD with the efficiency of explicit feature mapping and the adaptability of online learning. Our method employs Random Fourier Features~\citep{rahimi2007random} to explicitly lift time series data into a finite-dimensional reproducing kernel Hilbert space, avoiding the computational pitfalls of implicit kernel methods while capturing nonlinear dynamics. By coupling this with Sherman-Morrison-based online updates~\citep{zhang2019online}, WORK-DMD maintains fixed computational complexity per update while adapting continuously to streaming data.

 WORK-DMD requires only the current windowed snapshot to generate predictions, unlike deep learning methods that need extensive training datasets and multiple passes through historical data. This property makes it particularly suitable for applications with limited historical data or scenarios where rapid deployment is essential. The method naturally handles non-stationarity through its rolling window mechanism and online updates, discarding outdated information while incorporating new patterns without suffering from catastrophic forgetting.

Our contributions are threefold: (1) We present the first integration of Random Fourier Features with online DMD, enabling efficient kernel-based nonlinear modeling for streaming time series while maintaining interpretable DMD eigenvalue structure. (2) We develop a decoding mechanism that maps feature space predictions back to physical coordinates, ensuring practical applicability without compromising computational efficiency. (3) We provide a comprehensive evaluation against the latest online forecasting methods in multiple benchmark datasets, demonstrating competitive or superior performance with substantially lower data requirements and a single-pass learning process. The remainder of this paper details our methodology, presents extensive experimental validation, and discusses the implications of our findings for practical streaming forecasting applications. 

\section{Mathematical Background}
\label{sec: mathback}

\subsection{Space–time transformation of the data}

\noindent
Consider a multivariate time series
\[
\{\mathbf{x}_t\}_{t=1}^T,\quad
\mathbf{x}_t = \bigl(x_t^{(1)},\,\dots,\,x_t^{(p)}\bigr)^{\!\top}\in\mathbb{R}^p,
\]
where \(p\) is the number of components or features. Given a window length \(w\in\mathbb{N}\), we form the windowed data matrix
\begin{equation}
\label{eq:window}
\mathbf{X}_t
=
\begin{bmatrix}
\mathbf{x}_{t-w+1} & \mathbf{x}_{t-w+2} & \cdots & \mathbf{x}_t
\end{bmatrix}
\in\mathbb{R}^{p\times w}.
\end{equation}

\noindent
Its \(j\)th row
\(\mathbf{X}_t^{(j)}\in\mathbb{R}^{1\times w}\)
collects component \(j\) over the last \(w\) time steps:
\begin{equation}
\label{eq:Row_J}
\mathbf{X}_t^{(j)}
=
\bigl[x_{t-w+1}^{(j)},\,\dots,\,x_t^{(j)}\bigr].
\end{equation}

\noindent
Next, we choose an autoregressive depth \(d\le w\). For each row $\mathbf{X}_t^{(j)}, \ j=1,\dots,p$, we then construct the univariate Hankel matrix as
\begin{equation}
\label{eq:unihankel}
\boldsymbol{\mathcal{X}}_t^{(j)}
=
\begin{bmatrix}
x_{t-w+1}^{(j)} & \cdots & x_{t-d+1}^{(j)}\\
x_{t-w+2}^{(j)} & \cdots & x_{t-d+2}^{(j)}\\
\vdots          & \ddots & \vdots\\
x_{t-w+d}^{(j)} & \cdots & x_t^{(j)}
\end{bmatrix}
\;\in\;\mathbb{R}^{d\times (w-d+1)}.
\end{equation}

\noindent
Finally, we stack these \(p\) blocks to form the complete block–Hankel embedding:
\begin{equation}
\label{eq:blockhankel}
\boldsymbol{\mathcal{X}}_t
=
\begin{bmatrix}
\boldsymbol{\mathcal{X}}_t^{(1)}\\
\boldsymbol{\mathcal{X}}_t^{(2)}\\
\vdots\\
\boldsymbol{\mathcal{X}}_t^{(p)}
\end{bmatrix}
\in\mathbb{R}^{pd \times (w-d+1)}.
\end{equation}

\noindent
By employing a block–Hankel embedding to capture multi‐step temporal correlations, we facilitate the discovery of sparse, robust eigen-analysis that is required for dynamic mode decomposition ~\citep{brunton2017chaos,leclainche2017hodmd}.

\subsection{Dynamic Mode Decomposition}

\noindent
Fix \(m = w - d\).  From the current block–Hankel matrix in (\ref{eq:blockhankel}), we extract
\begin{equation}
\label{eq:dmd-data}
\begin{aligned}
\mathbf{X}
&=\bigl[\mathbf{x}_{t,1},\,\dots,\,\mathbf{x}_{t,m}\bigr],\\
\mathbf{Y}
&=\bigl[\mathbf{x}_{t,2},\,\dots,\,\mathbf{x}_{t,m+1}\bigr].
\end{aligned}
\end{equation}

\noindent
Here, \(\mathbf{x}_{t,i}\) denotes the \(i\)th column of the block–Hankel matrix \(\bm{\mathcal{X}}_t\). Consequently, the columns of \(\mathbf{Y}\) are each shifted one time step ahead of those in \(\mathbf{X}\). Given snapshot matrices \(\mathbf{X},\mathbf{Y}\) from \eqref{eq:dmd-data}, we seek an operator \(\mathbf{A}\) satisfying
\[
\mathbf{Y}\;\approx\;\mathbf{A}\,\mathbf{X}.
\]
The minimum‐norm least‐squares solution defines the standard DMD operator:
\begin{equation}
\label{eq:dmd-A-DMD}
\mathbf{A}^{\mathrm{DMD}}
\;=\;
\mathbf{Y}\,\mathbf{X}^{\dagger},
\end{equation} 
where \(\dagger\) denotes the Moore–Penrose pseudoinverse.  Further, we perform a rank-\(r\) truncated singular value decomposition  
\begin{equation}
\label{eq:dmd-svd-trunc}
\mathbf{X}
= \mathbf{Q}_X\,\boldsymbol{\Sigma}_X\,\mathbf{V}_X^*,
\end{equation}  
with \(\mathbf{Q}_X\in\mathbb{C}^{pd\times r}\), \(\boldsymbol{\Sigma}_X\in\mathbb{C}^{r\times r}\), \(\mathbf{V}_X\in\mathbb{C}^{m\times r}\), and \((\cdot)^*\) the conjugate transpose.  Projecting \(\mathbf{A}^{\mathrm{DMD}}\) into this reduced basis yields  
\begin{equation}
\label{eq:dmd-A-tilde}
\widetilde{\mathbf{A}}
= \mathbf{Q}_X^*\,\mathbf{A}^{\mathrm{DMD}}\,\mathbf{Q}_X
= \mathbf{Q}_X^*\,\mathbf{Y}\,\mathbf{V}_X\,\boldsymbol{\Sigma}_X^{-1}.
\end{equation}  
Its eigen-decomposition  
\[
\widetilde{\mathbf{A}}\,\mathbf{W} = \mathbf{W}\,\boldsymbol{\Lambda},\quad
\mathbf{W},\boldsymbol{\Lambda}\in\mathbb{C}^{r\times r},
\]
provides eigenvalues \(\boldsymbol{\Lambda}\) and eigenvectors \(\mathbf{W}\), from which the DMD modes are recovered as
\begin{equation}
\label{eq:dmd-modes-final}
\boldsymbol{\Phi}
= \mathbf{Q}_X\,\mathbf{W}
\;\in\;\mathbb{C}^{pd\times r}.
\end{equation}
Finally, the initial amplitudes are obtained from the last column of the snapshot matrix \(\mathbf{Y}\):
\begin{equation}
\label{eq:dmd-amplitudes}
\mathbf{b}_0 
= 
\boldsymbol{\Phi}^{\dagger}\,\mathbf{x}_{t,m+1}\,.
\end{equation}
Forecasting \(k\) steps into the future in delay‐coordinate space then proceeds by
\begin{equation}
\label{eq:dmd-forecast-future}
\widehat{\mathcal{X}}_k
= 
\boldsymbol{\Phi}\,\boldsymbol{\Lambda}^{\,k}\,\mathbf{b}_0\,.
\end{equation}
Selecting the appropriate \(p\) rows of \(\widehat{\mathcal{X}}_k\) yields the \(p\)-dimensional forecast \(\widehat{\mathbf{x}}_{t+k}\).

\section{Methodology}\label{sec:Method}

\begin{figure}[htbp]
    \centering
    \includegraphics[width=0.90\textwidth]{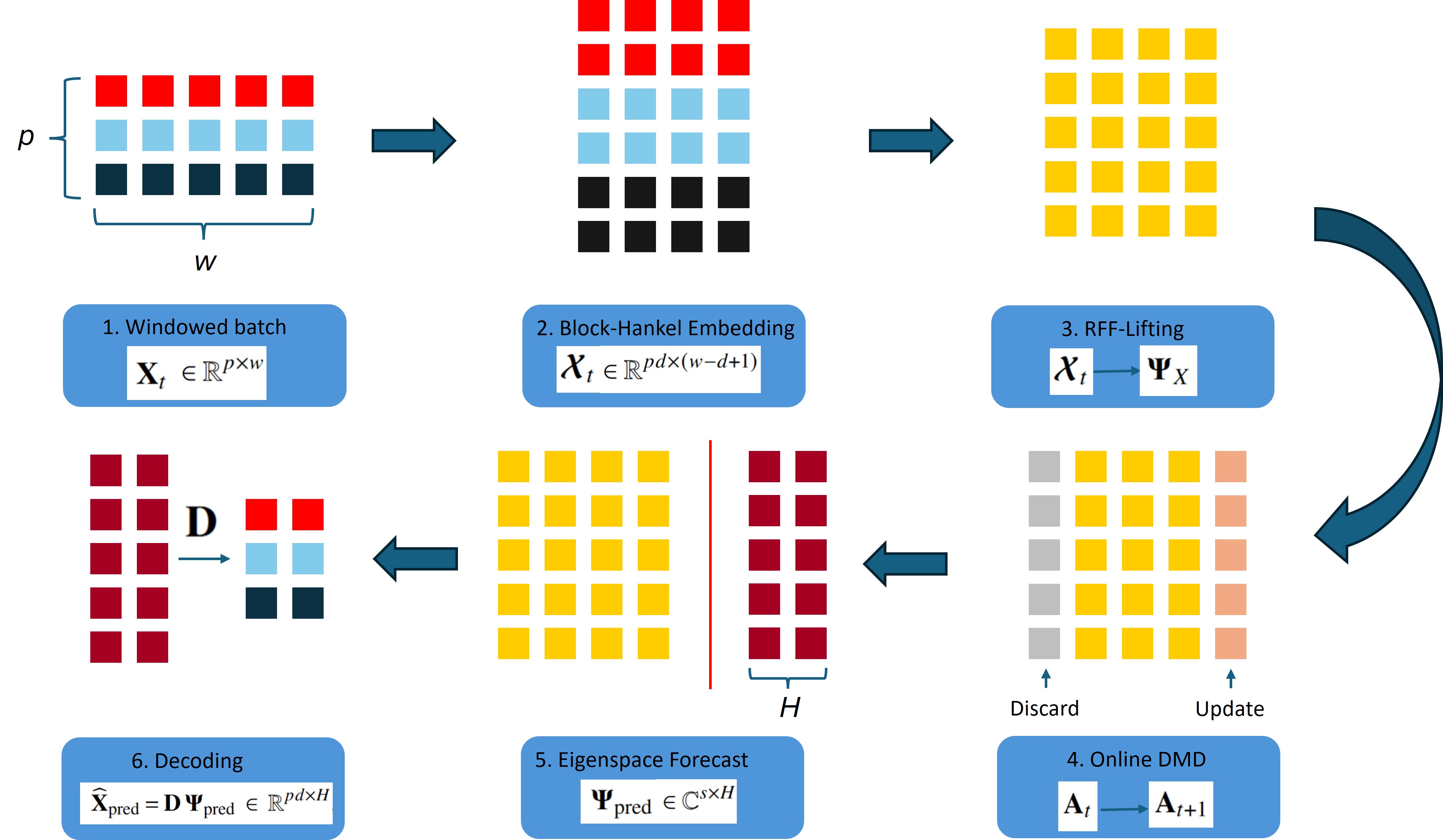}
    \caption{WORK-DMD methodology pipeline. The method processes multivariate time series data through the following steps: (1) Input multivariate windowed time series $\mathbf{X}_t \in\mathbb{R}^{p\times w}$ with $p$ features, (2) Construction of block-Hankel embedding $\boldsymbol{\mathcal{X}}_t \in \mathbb{R}^{pd \times m}$ to capture temporal correlations across multiple lag orders, (3) Random Fourier Feature (RFF) lifting to transform data into kernel feature space $\boldsymbol{\Psi}_X, \boldsymbol{\Psi}_Y \in \mathbb{R}^{s \times m}$ using Gaussian kernel approximation, (4) Online Dynamic Mode Decomposition update using Sherman-Morrison formulation to maintain Koopman operator $\mathbf{A}_{t+1} \in \mathbb{R}^{s \times s}$ with streaming data by discarding old snapshots (gray) and incorporating new observations (orange), (5) Feature-space forecasting through eigendecomposition in compressed POD basis to generate predictions $\boldsymbol{\Psi}_{\text{pred}} \in \mathbb{C}^{s \times H}$ (red divider separates current features from future predictions), and (6) Decoding via matrix $\mathbf{D}$ to transform feature-space predictions back to physical coordinates $\hat{\mathbf{x}}_{t+h} \in \mathbb{R}^p$. Color coding maintains feature identity throughout the pipeline: original time series features (red, teal, blue) are preserved through Hankel embedding, transformed to kernel features (yellow), processed through online updates, and decoded back to multivariate predictions.}
    \label{fig:work_dmd_methodology}
\end{figure}

We now introduce our new
method, which we call Windowed Online Random Kernel-Dynamic Mode Decomposition, or WORK-DMD for short. Our approach consists of three main steps.  
\textbf{Step 1} explicitly lifts the data to approximate a Gaussian kernel using Random Fourier Features (RFF)~\citep{rahimi2007random}, to enable nonlinear modeling while keeping the feature dimension manageable.  
\textbf{Step 2} performs an online kernel-DMD update each time a new snapshot arrives and produces forecasts.  
\textbf{Step 3} decodes those forecasts through the current DMD modes and eigenvalues to obtain the final prediction. A general overview of the entire method is shown in Figure \ref{fig:work_dmd_methodology}.

\subsection{Random-feature lifting}

Kernel methods usually exploit the kernel trick to sidestep high-dimensional computations, as shown in kernel DMD~\citep{kevrekidis2016kernel}.  In our setting, the product of series dimension \(p\), Hankel depth \(d\), and window size \(w\) can be large, so direct kernel arithmetic (cf.\ Section ~\ref{sec: mathback}) can be impractical.

To control dimensionality, we \emph{explicitly} lift each block–Hankel snapshot from \eqref{eq:blockhankel} into an \(s\)-dimensional space with RFF.  For a Gaussian kernel of bandwidth \(\gamma\), the mapping is
\begin{equation}
\label{eq:RFF}
\psi(\mathbf{x})
=
\sqrt{\frac{2}{s}}
\Bigl[
\cos\bigl(\theta_i + \mathbf{z}_i^{\!\top}\mathbf{x}\bigr)
\Bigr]_{i=1}^{s},
\end{equation}
where \(\theta_i\) are drawn uniformly from \([0,2\pi)\) and the random vectors \(\mathbf{z}_i\in\mathbb{R}^{pd}\) follow a multivariate normal distribution whose covariance is determined by \(\gamma\).  A detailed pseudocode for computing ~\eqref{eq:RFF} is given in~\citep{giannakis2023learning}. Applying \(\psi\) column-wise to the data blocks \(\mathbf{X}\) and \(\mathbf{Y}\) from \eqref{eq:dmd-data} yields the lifted snapshots
\begin{equation}
\label{eq:lifted-snapshots}
\bm{\Psi}_X = \psi\!\left(\mathbf{X}\right),
\quad
\bm{\Psi}_Y = \psi\!\left(\mathbf{Y}\right)
\in\mathbb{R}^{s\times m},
\end{equation}
which are then used in the subsequent online kernel-DMD step.

\subsection{Online Kernel DMD}

We adapt the update steps described in \citep{zhang2019online} for the online DMD problem. For the initial window size, we compute the equivalent DMD operator from (\ref{eq:dmd-A-DMD}) with 
\begin{equation}
\label{eq:batch-init}
\mathbf{P}_t
=
\bigl(\bm{\Psi}_X\bm{\Psi}_X^{\!\top}+\varepsilon\mathbf{I}\bigr)^{-1},
\qquad
\mathbf{A}_t
=
\bm{\Psi}_Y\bm{\Psi}_X^{\!\top}\mathbf{P}_t\, .
\end{equation}
In practice, $\varepsilon\mathbf I$ is used to avoid numerical problems during the inverse operation; $\varepsilon$ taken to be $10^{-6}\lVert \bm{\Psi}_X\bm{\Psi}_X^{\!\top} \rVert_2$. We use the Sherman-Morrison formulation \citep{zhang2019online} to update (\ref{eq:batch-init}), which ultimately contains the components that allow forecasting. 

Consider a rolling window of $\mathbf{X}, \mathbf{Y}$, where the oldest snapshot columns $\mathbf{x}_{t,1}, \mathbf{x}_{t,2}$ are 
discarded and new snapshots $\mathbf{x}_{t,m+1}, \mathbf{x}_{t,m+2}$ are introduced. We define the following matrices:
\begin{equation}
\label{eq:U_V_C}
\begin{aligned}
\mathbf{U} &= \bigl[\psi(\mathbf{x}_{t,1}) \;\;\; \psi(\mathbf{x}_{t,m+1})\bigr],\\[4pt]
\mathbf{V} &= \bigl[\psi(\mathbf{x}_{t,2}) \;\;\; \psi(\mathbf{x}_{t,m+2})\bigr],\\[4pt]
\mathbf{C} &=
\begin{bmatrix}
-1 & 0\\
 0 & 1
\end{bmatrix}.
\end{aligned}
\end{equation}
We also define the auxiliary matrix $\bm{\Gamma}$ as
\begin{equation}
\label{eq:gamma}
\bm{\Gamma} =
\bigl(\mathbf{C}^{-1}+\mathbf{U}^{\!\top}\mathbf{P}_t\mathbf{U}\bigr)^{-1},
\end{equation}
which captures how much the current $\mathbf{P}_t$ matrix aligns with the directions of the new and old snapshots. The set of Sherman-Morrison equations \citep{zhang2019online} is then used to update the following:
\begin{align}
\mathbf{P}_{t+1} &=
\mathbf{P}_t\;-\;\mathbf{P}_t\mathbf{U}\,\bm{\Gamma}\,\mathbf{U}^{\!\top}\mathbf{P}_t,
\label{eq:P_update}\\[4pt]
\mathbf{A}_{t+1} &=
\mathbf{A}_t\;+\;\bigl(\mathbf{V}-\mathbf{A}_t\mathbf{U}\bigr)
\,\bm{\Gamma}\,\mathbf{U}^{\!\top}\mathbf{P}_t.
\label{eq:A_update}
\end{align}
The intuitive interpretation of the above steps is that the term $\bigl(\mathbf{V}-\mathbf{A}_t\mathbf{U}\bigr)$ in equation \eqref{eq:A_update}, scaled by $\bm{\Gamma}$, can be considered a prediction error of the current model $\mathbf{A}_t$. The above formulae are meant to be used in a recursive manner.

\subsection{Feature-space forecasting}\label{sec:feature-forecast}

After each online update, we have the current Koopman operator  
\(\mathbf{A}_{t+1}\in\mathbb{C}^{s \times s}\) and its rank-\(r\) POD basis  
\(\mathbf{Q}_{r}\in\mathbb{R}^{s \times r}\), obtained from the refreshed feature matrix \(\bm{\Psi}_X\).  
Forecasting then proceeds in three algebraic steps.

\paragraph{(i) Projection onto the POD subspace.}
\begin{equation}
\label{eq:K_reduced}
\mathbf{K}_{t+1}
=
\mathbf{Q}_{r}^{\!\top}\,\mathbf{A}_{t+1}\,\mathbf{Q}_{r}
\ \in\;\mathbb{R}^{r\times r}.
\end{equation}

\paragraph{(ii) Eigendecomposition in the reduced space.}
\begin{equation}
\label{eq:eig_reduced}
\begin{aligned}
\mathbf{K}_{t+1}\,\mathbf{W}_{r} &= \mathbf{W}_{r}\,\boldsymbol{\Lambda}_{r},\\
\mathbf{W}_{r},\;\boldsymbol{\Lambda}_{r} \ &\in \ \mathbb{C}^{r\times r},\\
\boldsymbol{\Lambda}_{r} &= \operatorname{diag} \ \bigl(\lambda_{1},\dots,\lambda_{r}\bigr).
\end{aligned}
\end{equation}

\paragraph{(iii) Forecast in the compressed basis.}
\begin{align}
\mathbf{b}_{0} &=
\mathbf{W}_{r}^{-1}\,\mathbf{Q}_{r}^{\!\top}\,
\psi\!\bigl(\mathbf{x}_{t,m+1}\bigr),
\label{eq:b0}\\[4pt]
\mathbf{E} &=
\bigl[\lambda_{i}^{\,h}\bigr]_{\,i=1,\,h=0}^{\,r,\,H-1} \
\in \ \mathbb{C}^{r\times H},
\label{eq:Vandermonde}\\[4pt]
\bm{\Psi}_{\text{pred}} &=
\mathbf{Q}_{r}\,\mathbf{W}_{r}\,
\bigl(\mathbf{b}_{0}\odot\mathbf{E}\bigr)
\ \in \ \mathbb{C}^{s\times H},
\label{eq:psi_pred}
\end{align}
where \(\odot\) denotes element-wise multiplication.  
Equation~\eqref{eq:psi_pred} is the compressed analog of the full-space forecast.  
Its columns give the feature-space trajectories  
\(\psi\!\bigl(\widehat{\mathbf{x}}_{t,m+1+h}\bigr)\) for (time) horizon \(h=1,\dots,H\), with \(\mathbf{x}_{t,m+1}\) being the most recent snapshot after the update. The matrix \(\mathbf{E}\) is a Vandermonde-like structure encoding the temporal evolution of each eigenvalue across all forecast horizons \citep{tirunagari2017dynamic}, where entry \(E_{ih} = \lambda_i^h\) captures how the \(i\)-th mode propagates \(h\) steps forward in time.

\subsection{Decoding to the physical space}
\label{sec:decoding}

To convert feature–space forecasts back to the original coordinates we build a
\emph{decoder} matrix from the current lifted snapshot pair.  
Let \(\bm{\Psi}_X\in\mathbb{R}^{s\times m}\) and 
\(\mathbf{X}\in\mathbb{R}^{pd\times m}\) denote, respectively,
the lifted and physical Hankel blocks at time step \(t+1\).
The decoder is obtained by a one–time ridge‐free least–squares fit:
\begin{equation}
\label{eq:decoder-def}
\mathbf{D}
=
\mathbf{X}\,
\bm{\Psi}_X^{\dagger}
\;\in\;\mathbb{R}^{pd\times s}.
\end{equation}
Applying \(\mathbf{D}\) to the feature–space trajectory
\(\bm{\Psi}_{\text{pred}}\) from \eqref{eq:psi_pred} yields the decoded forecast matrix
\begin{equation}
\label{eq:decoded-forecast}
\widehat{\mathbf{X}}_{\text{pred}}
=
\mathbf{D}\,
\bm{\Psi}_{\text{pred}}
\;\in\;\mathbb{R}^{pd\times H},
\end{equation}
whose columns provide the predicted snapshots
\(\widehat{\mathbf{x}}_{t,m+1+h}\), for horizon \(h=1,\dots,H\), in the original data space.

\section{Results}\label{sec:Results}

This section describes the experimental data sets and the online forecasting results.

\subsection{Datasets}

The datasets used in this study have typically served as benchmarks to generalize time series forecasting performance of a model, particularly with state-of-the-art methods such as deep learning models \citep{li2024deep}. We use it to compare to other recent models who have their own methods for the online time series forecasting problem. Electricity Transformer Temperature (ETT) \citep{yue2022ts2vec} captures load and oil temperature data of power transformers at 15-minute intervals from July 2016 to July 2018. It is composed of hourly granularity data (ETTh2) and 15-minute granularity data (ETTm1). Traffic \citep{traffic} contains the hourly road occupancy rates measured by 862 sensors in the San Francisco Bay area freeways from January 2015 to December 2016. The Weather (WTH) \citep{WTH} dataset contains hourly records of 11 climate features from almost 1,600 locations across the United States.

\subsection{Implementation details}
We follow the implementation details used in the FSNet method \citep{pham2022learning}, where the data are split into warm-up and online testing phases by a 25:75 ratio. However, since our method is not a traditional machine/deep learning model, we use the warm-up phase to tune our hyperparameters. We perform hyperparameter tuning for $\{r,d,\gamma,s\}$ via random search with 3-fold rolling cross-validation on the warm-up data, 
independently for each dataset. In addition, the mean and standard deviation are calculated in the warm-up phase to normalize the online testing samples. For all the benchmark datasets, we vary the forecasting window as \(H \in \{1,24,48\}\). We evaluate forecasting performance using Mean Squared Error (MSE) and Mean Absolute Error (MAE) as our primary error metrics. All experiments were conducted on an Nvidia GeForce RTX 3070 GPU with 8GB of RAM using PyTorch.

The optimized hyperparameter configurations for WORK-DMD show consistent values across all the datasets for gamma ($\gamma = 1 \times 10^{-4}$) and auto-regressive depth ($d = 30$), while window sizes and RFF dimensions are adapted to dataset characteristics. Specifically, ETTh2 and WTH utilize larger window sizes ($w = 120$) with RFF dimensions of $s = 1024$ and $512$, respectively, while ETTm1 and Traffic employ smaller windows ($w = 60$) with RFF dimensions of $s = 256$ and $512$, respectively. The rank parameters are determined by the column rank available in each experiment. Additional sensitivity studies that examine the impact of hyperparameter variations on performance are provided in Appendix \ref{sec:sensitivity}.

\subsection{Baseline of Adaptation Methods}
We compare our method against five recent online forecasting approaches: \textbf{OnlineTCN}~\citep{LeGuennec2016TCN,Bai2018TCN}, an adaptation of Temporal Convolutional Networks for incremental updates; \textbf{DER++}~\citep{buzzega2020dark}, a continual learning method mitigating catastrophic forgetting through replay and regularization; \textbf{Informer}~\citep{Zhou2021Informer}, a transformer model with Prob-Sparse attention for efficient long-sequence prediction; \textbf{FSNet}~\citep{pham2022learning}, a frequency-spectral network capturing multi-scale dependencies with fast and slow learning; and \textbf{OneNet}~\citep{zhang2023onenet}, a unified framework dynamically ensembling models for online forecasting under concept drift. These baselines span diverse architectures including transformer-based, convolutional, spectral, and continual learning, providing a balanced reference for evaluating online forecasting performance.

\subsection{Online Forecasting Results}

The forecasting results in Table~\ref{tab:ett_results} show that the proposed WORK-DMD method achieves competitive or superior accuracy compared to established baselines across diverse datasets and prediction horizons. Notably, WORK-DMD consistently achieves the lowest error metrics for short-term forecasts ($H=1$), where both MSE and MAE improvements are the most pronounced. Across all datasets, WORK-DMD matches or outperforms deep learning models such as OneNet, FSNet, and Informer, demonstrating its effectiveness at capturing both short-term dynamics and longer-term dependencies compared to the baseline methods.

\begin{table}[ht]
\centering
\caption{Multivariate Forecasting results, averaged over all variables (horizon $H$).}
\label{tab:ett_results}
\begin{tabular}{@{} c c
  cc   
  cc   
  cc   
  cc   
  cc   
  cc   
@{}}
\toprule
Dataset & $H$
  & \multicolumn{2}{c}{OnlineTCN}
  & \multicolumn{2}{c}{DER++}
  & \multicolumn{2}{c}{Informer}
  & \multicolumn{2}{c}{FSNet}
  & \multicolumn{2}{c}{OneNet}
  & \multicolumn{2}{c}{WORK-DMD} \\
 & 
  & MSE    & MAE
  & MSE    & MAE
  & MSE    & MAE
  & MSE    & MAE
  & MSE    & MAE
  & MSE    & MAE \\
\midrule
ETTh2 & 1  
  & 0.502  & 0.436
  & 0.508  & 0.375
  & 7.571  & 0.850
  & 0.466  & 0.368
  & 0.380  & 0.348   
  & \textbf{0.289} & \textbf{0.334} \\
      & 24 
  & 0.830  & 0.547
  & 0.828  & 0.540
  & 4.629  & 0.668
  & 0.687  & 0.467
  & 0.532  & \textbf{0.407}  
  & \textbf{0.480}  & 0.453 \\
      & 48 
  & 1.183  & 0.589
  & 1.157  & 0.577
  & 5.692  & 0.752
  & 0.846  & 0.515
  & 0.609  & \textbf{0.436}   
  & \textbf{0.603}  & 0.516 \\
\addlinespace
ETTm1 & 1  
  & 0.214  & 0.085
  & 0.083  & 0.192
  & 0.456  & 0.512
  & 0.085  & 0.191
  & 0.082  & 0.187
  & \textbf{0.004} & \textbf{0.102} \\
      & 24 
  & 0.258  & 0.381
  & 0.196  & 0.326
  & 0.478  & 0.525
  & 0.115  & 0.249
  & \textbf{0.098} & 0.225   
  & 0.102   & \textbf{0.200} \\
      & 48 
  & 0.283  & 0.403
  & 0.208  & 0.340
  & 0.377  & 0.460
  & 0.127  & 0.263
  & \textbf{0.108} & \textbf{0.238}   
  & 0.121   & 0.314 \\
\addlinespace
Traffic & 1  
  & 0.315  & 0.283
  & 0.289  & \textbf{0.248}
  & 0.795  & 0.507
  & 0.288  & 0.253
  & -      & -
  & \textbf{0.275} & 0.265 \\
      & 24 
  & 0.452  & 0.363
  & 0.387  & 0.295
  & 1.267  & 0.750
  & 0.362  & \textbf{0.288}
  & -      & -   
  & \textbf{0.357} & 0.291 \\
\addlinespace
WTH & 1  
  & 0.206  & 0.276
  & 0.174  & 0.235
  & 0.426  & 0.458
  & 0.162  & 0.216
  & 0.156  & 0.201 
  & \textbf{0.149} & \textbf{0.197} \\
      & 24 
  & 0.308  & 0.367
  & 0.287  & 0.351
  & 0.370  & 0.417
  & 0.188  & 0.276
  & \textbf{0.175} & 0.255   
  & 0.220   & \textbf{0.253} \\
      & 48 
  & 0.302  & 0.362
  & 0.294  & 0.359
  & 0.367  & 0.419
  & 0.223  & 0.301
  & \textbf{0.200} & 0.279  
  & 0.261   & \textbf{0.275} \\
\bottomrule
\end{tabular}
\begin{tablenotes}
\small
\item Note: OneNet results for the Traffic dataset were not provided by the authors.
\end{tablenotes}
\end{table}

\begin{figure}[htbp]
    \centering
    \begin{subfigure}[b]{0.8\textwidth}
        \centering
        \includegraphics[width=\textwidth]{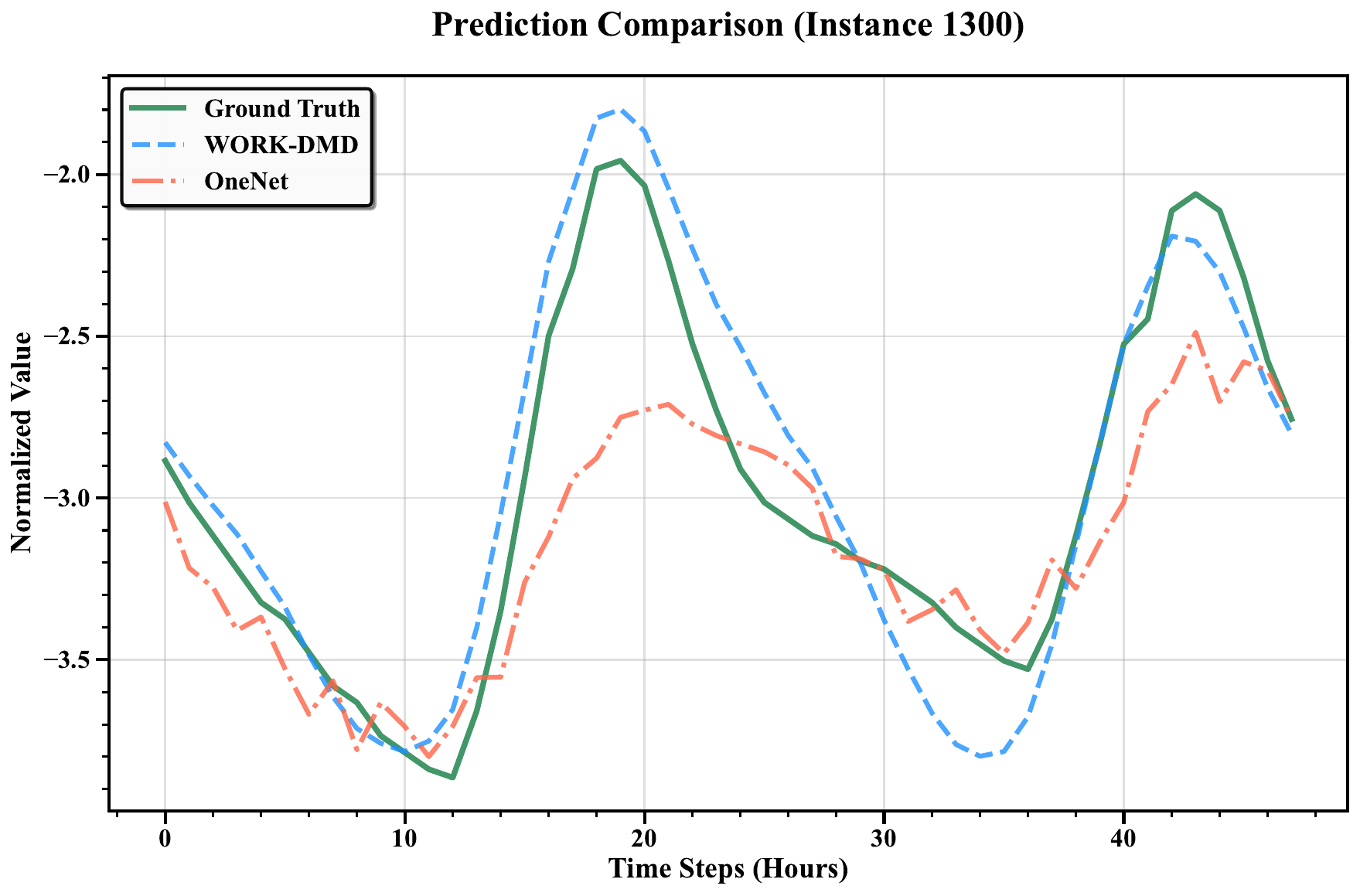}
        \caption{Instance 1300}
        \label{fig:ETT_H2_pred_1000}
    \end{subfigure}
    
    \vspace{0.4cm}
    
    \begin{subfigure}[b]{0.8\textwidth}
        \centering
        \includegraphics[width=\textwidth]{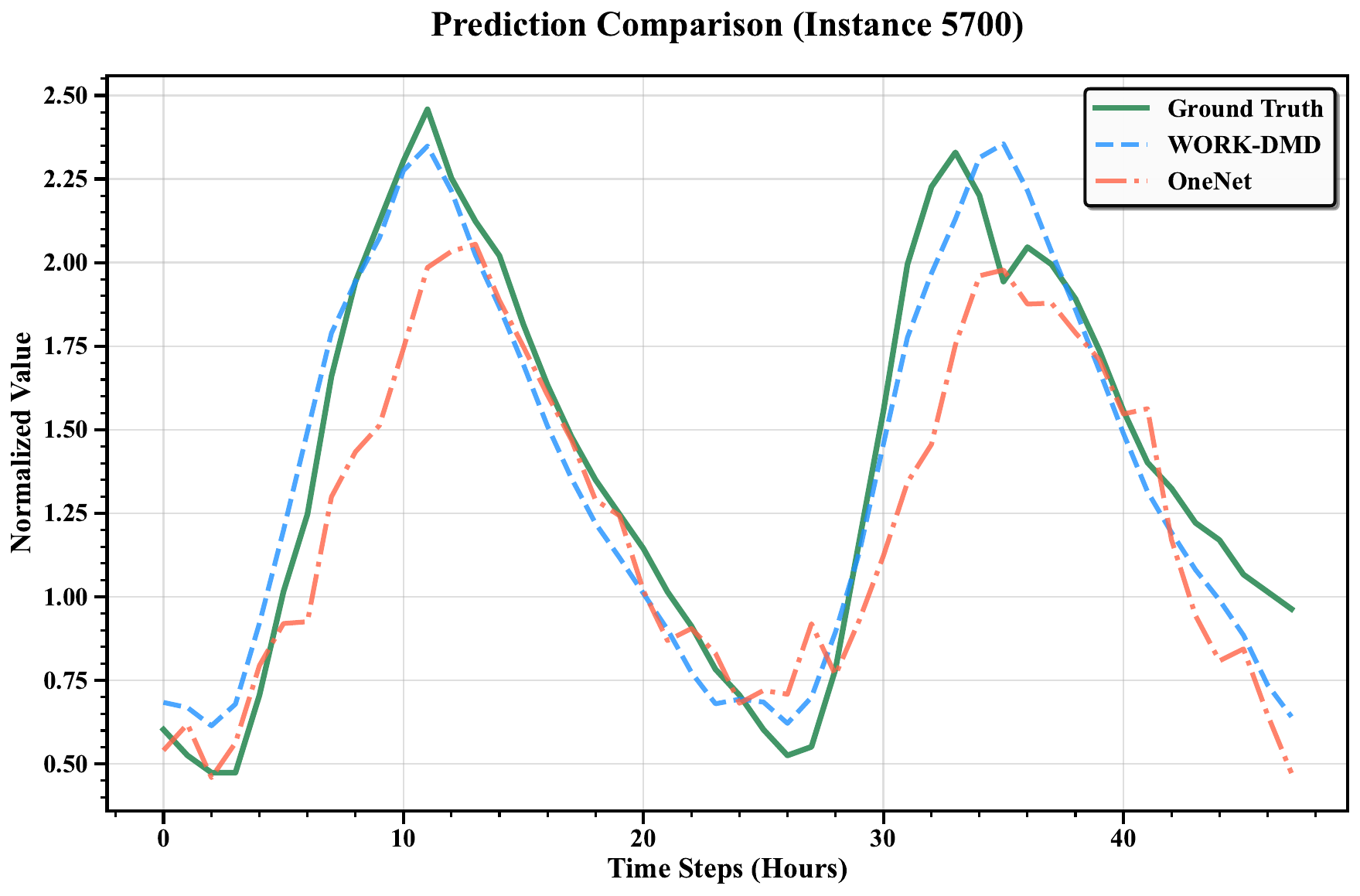}
        \caption{Instance 5700}
        \label{fig:ETT_H2_pred_5000}
    \end{subfigure}
    
    \caption{Time series forecasting comparison on ETTh2 for the Oil Temperature (OT) variable. The figures show 48-hour ahead predictions from WORK-DMD and OneNet against ground truth for two different instances: (a) Instance 1300 and (b) Instance 5700. In both instances, WORK-DMD demonstrates better adherence to the ground truth compared to OneNet.}
    \label{fig:ETT_H2_forecasting_comparison}
\end{figure}

\begin{figure}[htbp]
    \centering
    \includegraphics[width=0.8\columnwidth]{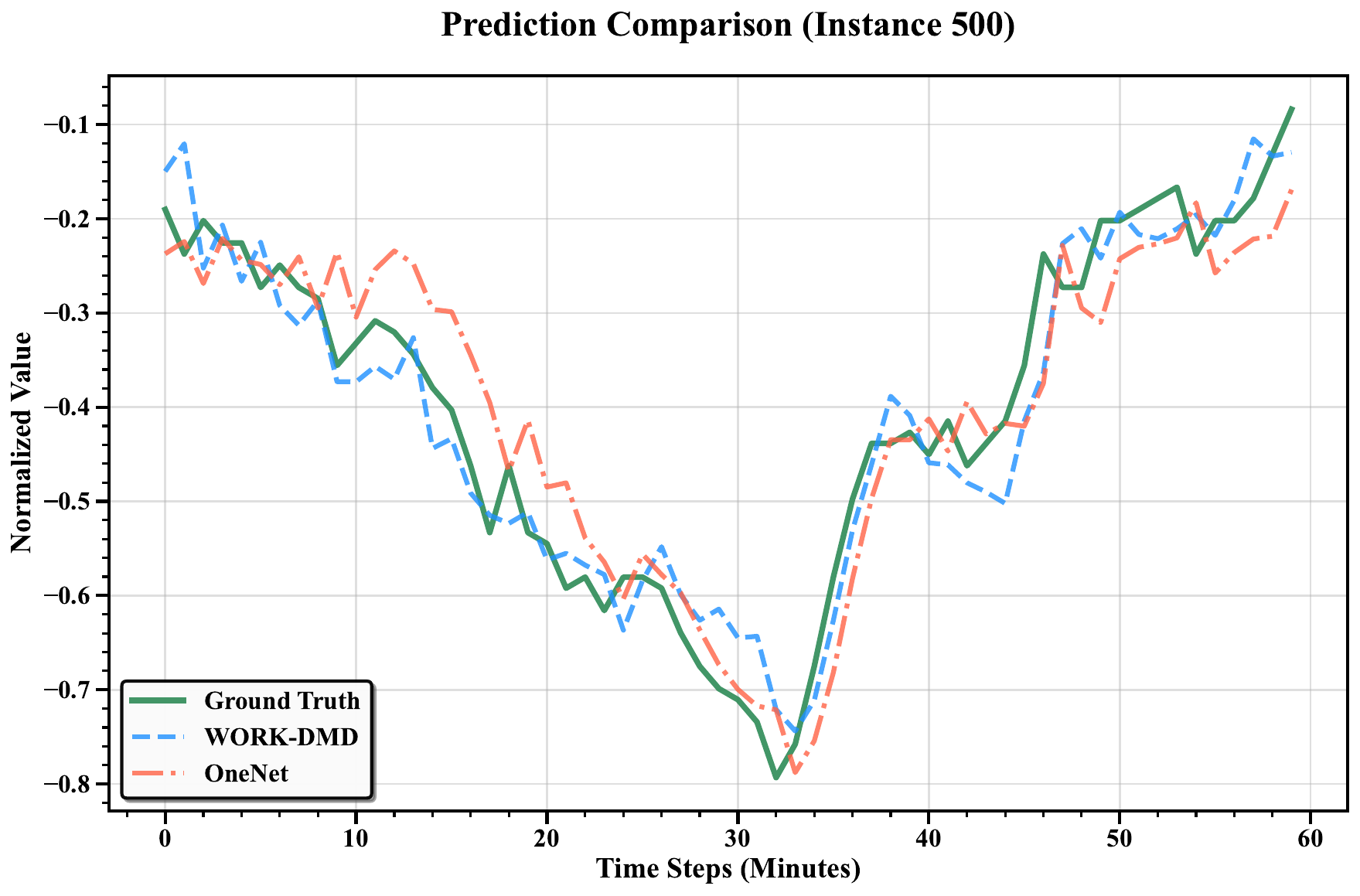}
    \caption{Time series forecasting comparison on ETTm1 for the Oil Temperature (OT) variable. The figure shows 1-step ahead predictions comparing WORK-DMD and OneNet against ground truth for instance 500. WORK-DMD exhibits marginally closer alignment to the ground truth, demonstrating detailed tracking of temporal patterns.}
    \label{fig:ETT_M1_forecasting_comparison}
\end{figure}

\begin{figure}[htbp]
    \centering
    \begin{subfigure}[b]{0.8\textwidth}
        \centering
        \includegraphics[width=\textwidth]{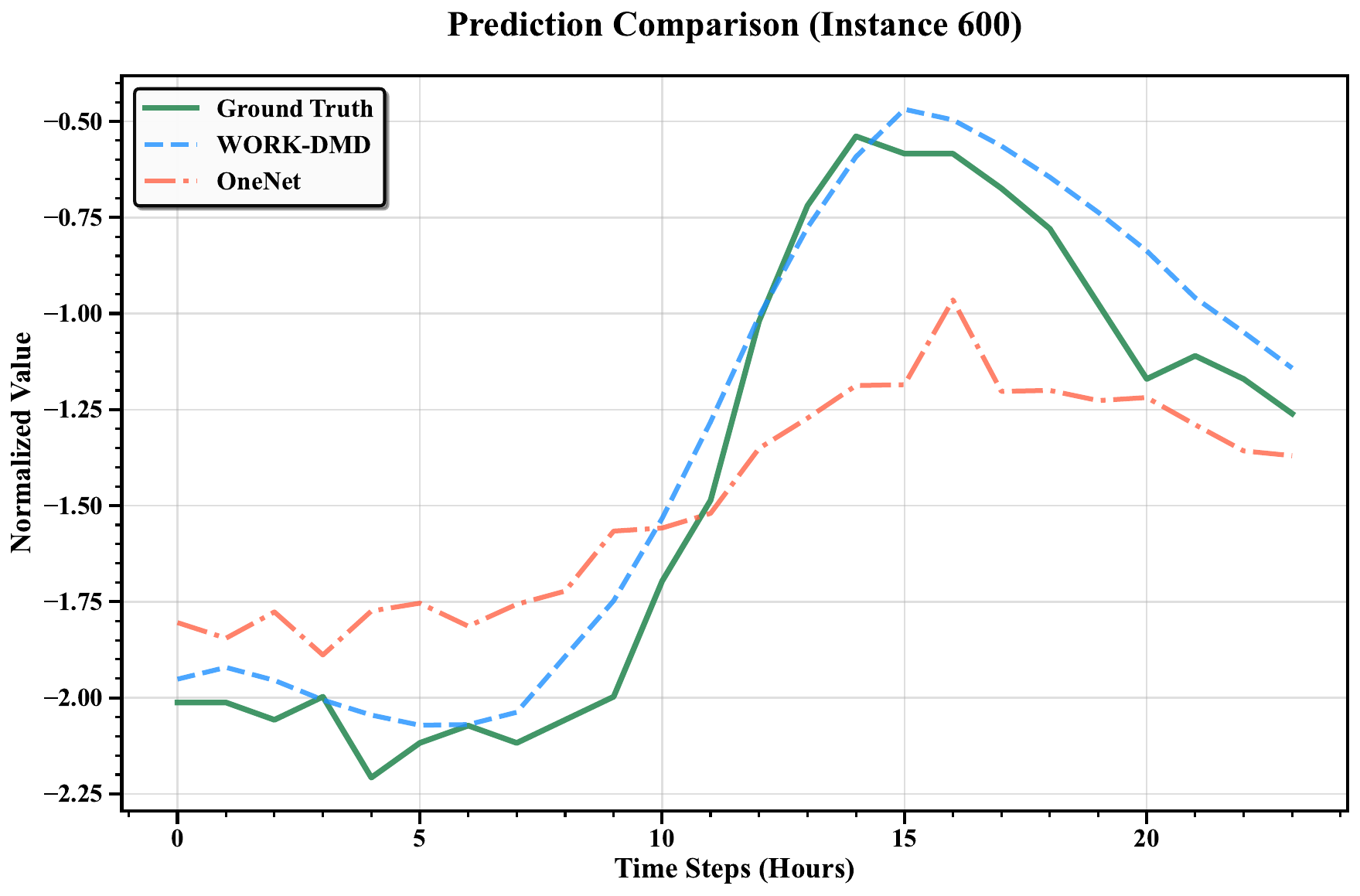}
        \caption{Instance 600}
        \label{fig:WTH_pred_1000}
    \end{subfigure}
    
    \vspace{0.4cm}
    
    \begin{subfigure}[b]{0.8\textwidth}
        \centering
        \includegraphics[width=\textwidth]{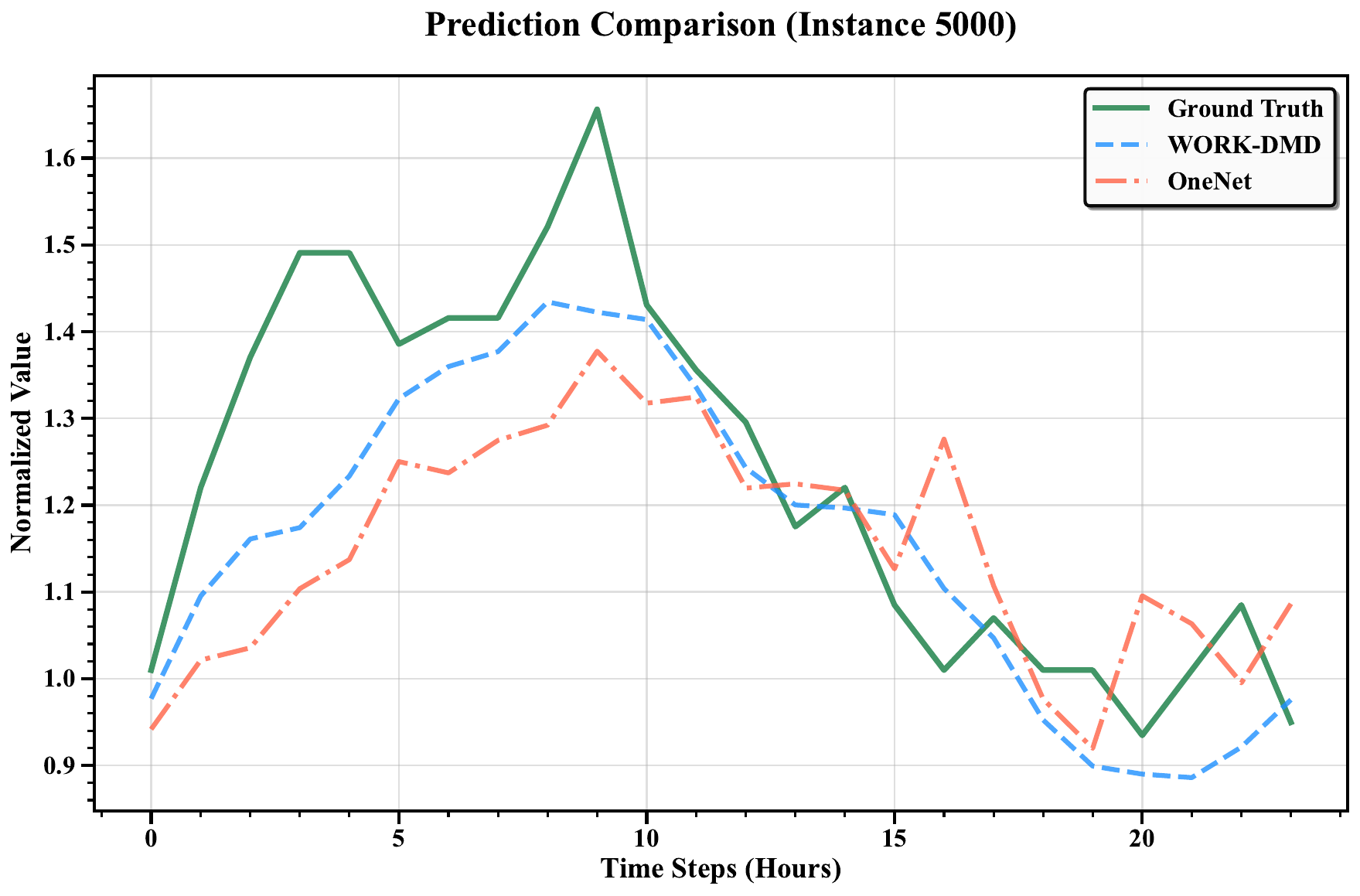}
        \caption{Instance 5000}
        \label{fig:WTH_pred_3500}
    \end{subfigure}
    
    \caption{Time series forecasting comparison on WTH dataset for the Wet Bulb Temperature variable. The figures show 24-hour ahead predictions from WORK-DMD and OneNet against ground truth for two different instances: (a) Instance 600 and (b) Instance 5000. In (a), WORK-DMD demonstrates better adherence to the ground truth compared to OneNet, while in (b) both methods exhibit comparable performance. Notably, WORK-DMD showcases adaptive capabilities by accurately forecasting across different temperature regimes, operating in predominantly negative values in (a) and transitioning to positive values in (b).}
\label{fig:WTH_forecasting_comparison}
\end{figure}

Further evidence of WORK-DMD's strong performance is seen in the following prediction plots, where we primarily compare with OneNet as it is the closest competitor. Figure \ref{fig:ETT_H2_forecasting_comparison} demonstrates that for longer $H = 48$ horizons, our method closely matches the cyclical behavior of the ETTh2 time series across two different instances. Furthermore, there is evidence that our method adapts to mean shifts in the time series, as shown by the model operating in negative value space in Figure \ref{fig:ETT_H2_forecasting_comparison}(a)
and positive value space in Figure \ref{fig:ETT_H2_forecasting_comparison}(b).
Figure \ref{fig:ETT_M1_forecasting_comparison} shows the consecutive single-step ahead forecasts against the ground truth, showcasing how well WORK-DMD adheres to the ground truth patterns for ETTm1. Figure \ref{fig:WTH_forecasting_comparison} illustrates two instances within the WTH time series for $H = 24$ forecasting of the Wet Bulb Temperature variable, showing that WORK-DMD can capture underlying dynamics despite erratic behavior at different stages of the time series.

\begin{figure}[htbp]
    \centering
    \begin{subfigure}[b]{0.8\textwidth}
        \centering
        \includegraphics[width=\textwidth]{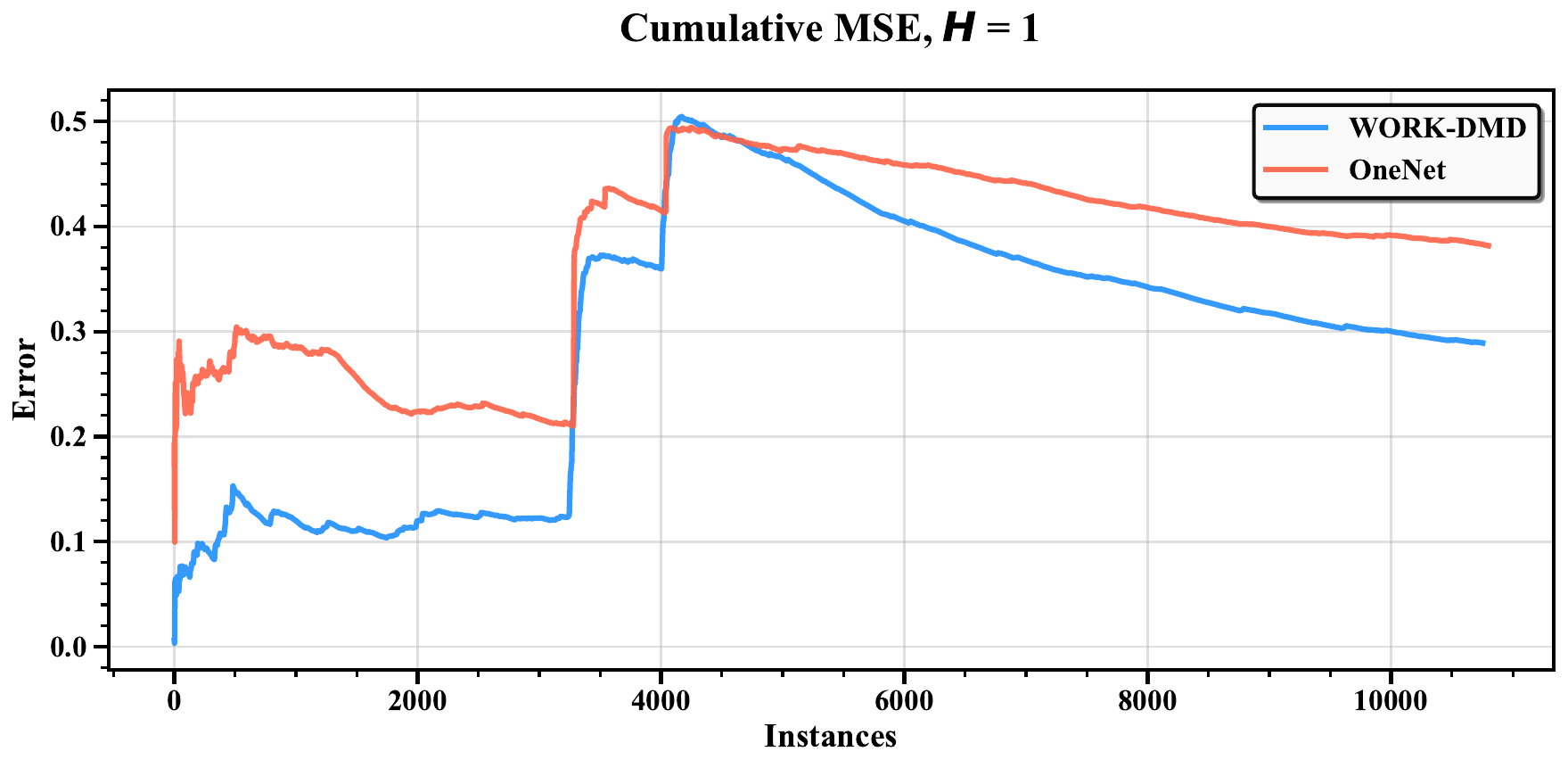}
        \caption{ETTh2 1-step ahead}
        \label{fig:ETTH2_cum_1}
    \end{subfigure}
    
    \vspace{0.4cm}
    
    \begin{subfigure}[b]{0.8\textwidth}
        \centering
        \includegraphics[width=\textwidth]{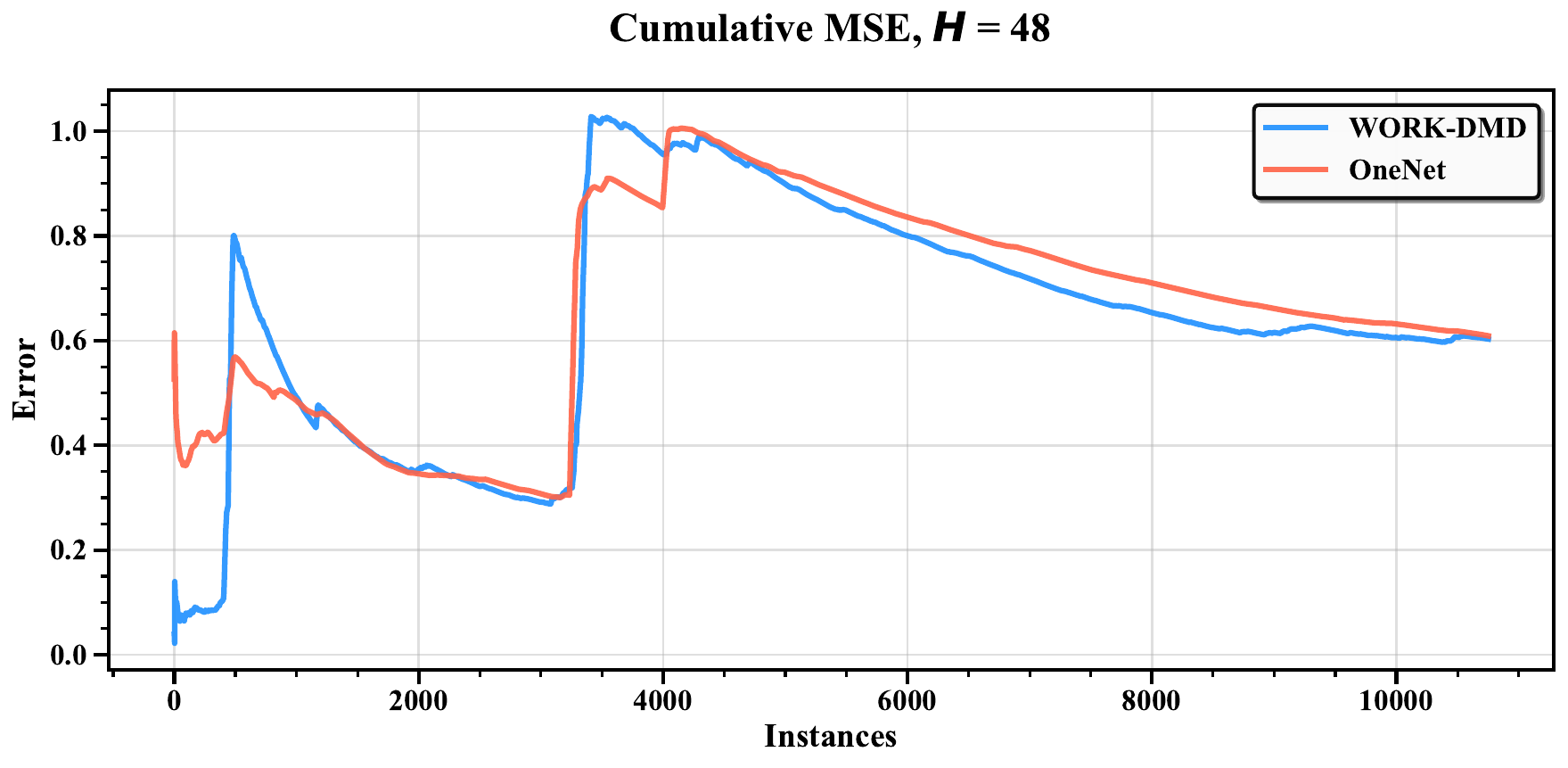}
        \caption{ETTh2 48-step ahead}
        \label{fig:ETTH_cum_48}
    \end{subfigure}
    
    \caption{Cumulative MSE comparison on ETTh2 dataset. The figures show cumulative mean squared error progression for (a) ETTh2 1-step ahead and (b) ETTh2 48-step ahead forecasting, comparing WORK-DMD and OneNet performance. In (a), WORK-DMD demonstrates faster error correction, resulting in lower cumulative error accumulation compared to OneNet. In (b), both methods exhibit comparable performance across the forecasting horizon.}
    \label{fig:ETT_cumulative_MSE}
\end{figure}

\begin{figure}[htbp]
    \centering
    \begin{subfigure}[b]{0.8\textwidth}
        \centering
        \includegraphics[width=\textwidth]{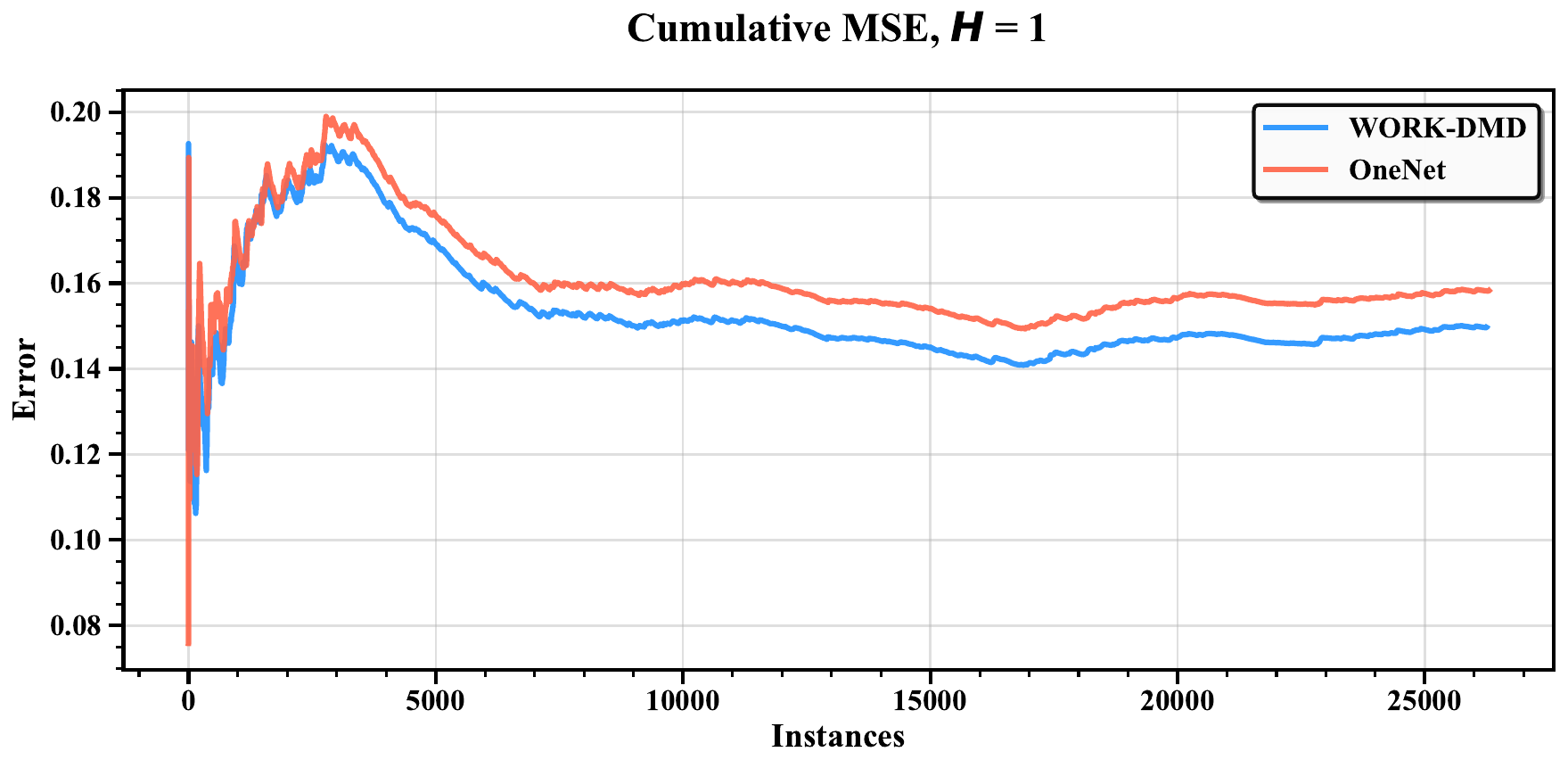}
        \caption{1-step ahead}
        \label{fig:WTH_cum_1}
    \end{subfigure}
    
    \vspace{0.4cm}
    
    \begin{subfigure}[b]{0.8\textwidth}
        \centering
        \includegraphics[width=\textwidth]{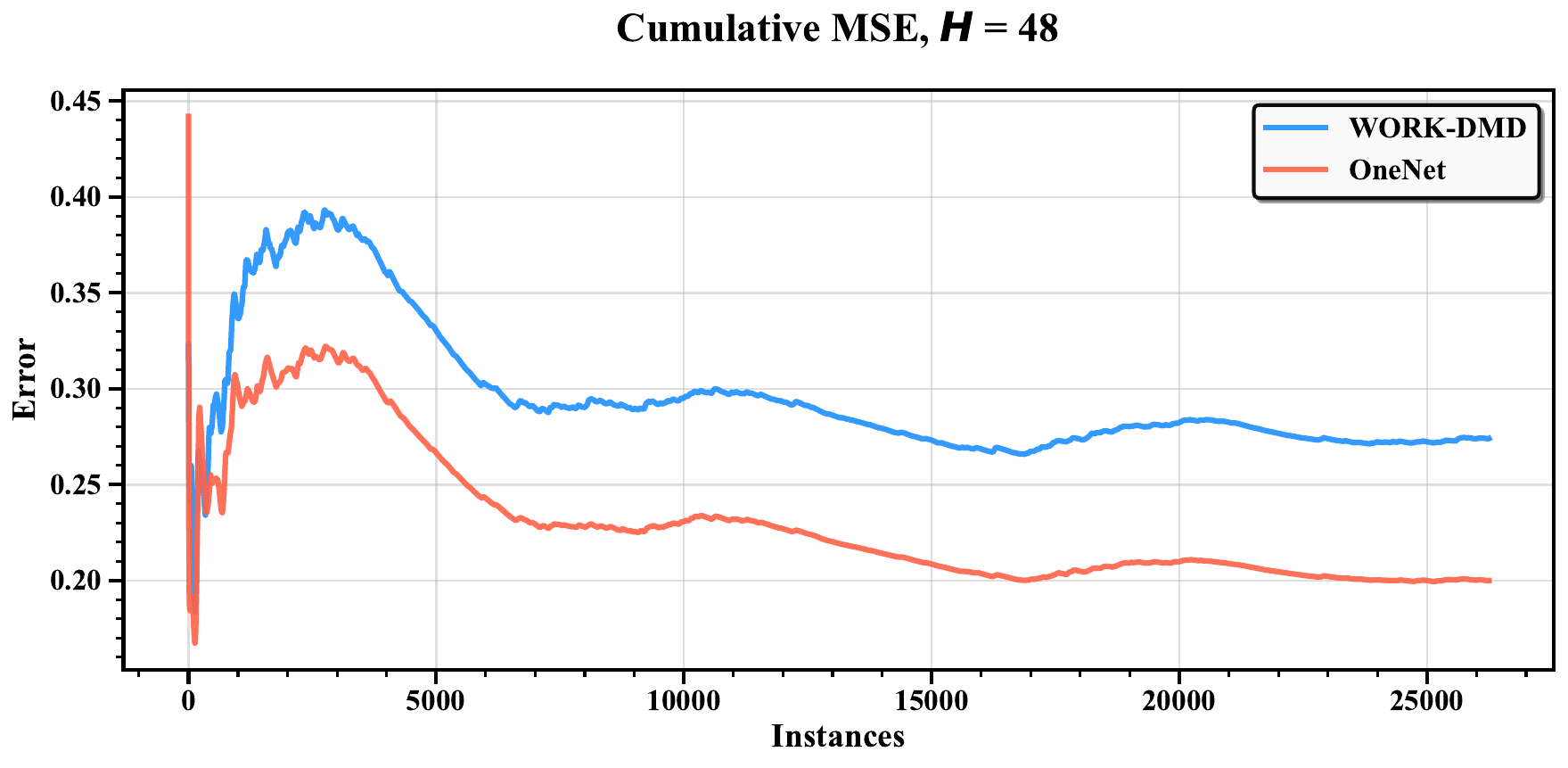}
        \caption{48-step ahead}
        \label{fig:WTH_cum_48}
    \end{subfigure}
    
    \caption{Cumulative MSE comparison on WTH dataset. The figures show cumulative mean squared error progression for (a) 1-step ahead and (b) 48-step ahead forecasting, comparing WORK-DMD and OneNet performance over time. In (a), WORK-DMD marginally outperforms OneNet with slightly lower cumulative error accumulation. In (b), OneNet demonstrates better performance, though WORK-DMD remains competitive and maintains close proximity throughout the forecasting horizon.}
    \label{fig:WTH_cumulative_MSE}
\end{figure}

Finally, to show the adaptive nature of our method, we report cumulative MSE plots for the ETTh2 and WTH datasets. Figure \ref{fig:ETT_cumulative_MSE} shows the cumulative MSE plots for $H = 1$ and $H = 48$ for ETTh2 compared to OneNet. Despite a drastic non-stationary shift in the ETTh2 time series, WORK-DMD is able to recover, sometimes at a faster rate than OneNet, eventually settling into a lower error score. Figure \ref{fig:WTH_cumulative_MSE} presents the cumulative MSE for $H = 1$ and $H = 48$ for the WTH dataset. We see that WORK-DMD, despite structural shifts in the time series, adapts to such changes due to our updating method discarding older snapshots that no longer represent the dynamics properly. Although OneNet achieves a lower error in Figure \ref{fig:WTH_cumulative_MSE}(b),
WORK-DMD remains within a competitive margin. Additional forecasting and cumulative error plots for the Traffic dataset are shown in Appendices \ref{sec: Traffic Forecast} and \ref{sec: app_error_plots}.

\subsection{Sample Efficiency}
A key advantage of WORK-DMD is its single-pass learning capability: each data point is processed only once, unlike deep learning methods that require multiple passes through the same data. To quantify this efficiency, we conduct a sequential online forecasting comparison between WORK-DMD, OneNet and FSNet using the Traffic dataset, tracking the total number of sample exposures each method requires during the training and online operation. Our experiment uses 100 traffic features with both methods starting from identical initial training data (60 time steps) and processing 100 streaming updates sequentially. WORK-DMD sees each sample once during training and once per online update, while OneNet/FSNet processes training data through 10 epochs and performs 1 epoch per online update using previous ground truth for supervision.

\begin{figure}[htbp]
    \centering
    \includegraphics[width=0.9\textwidth]{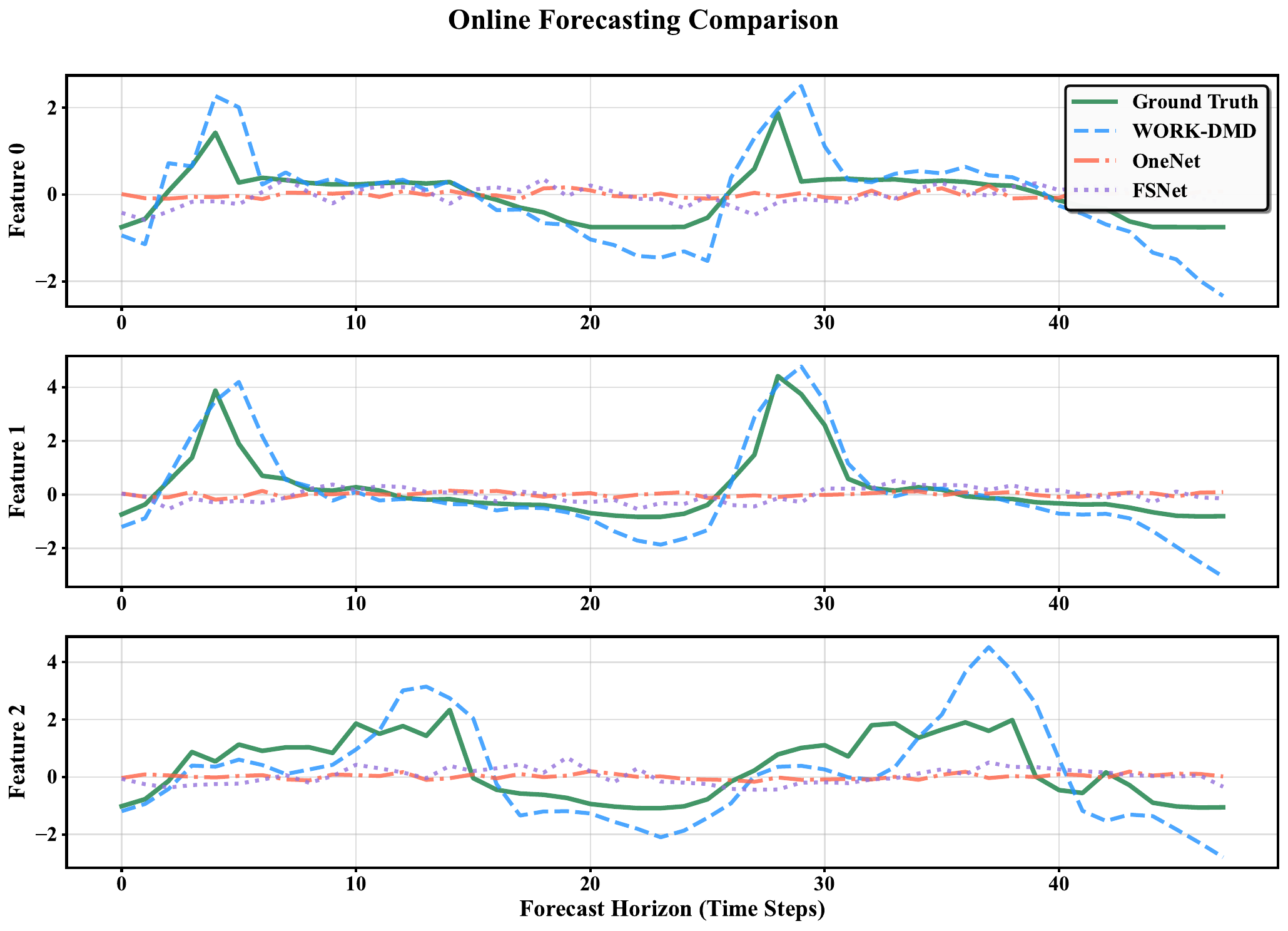}
    \caption{Multivariate online forecasting comparison on Traffic dataset for first three monitoring channels under limited training data. All methods start from identical initial training data and process streaming updates sequentially. WORK-DMD sees each sample once during training and once per update, while FSNet/OneNet requires multiple training epochs plus additional epochs per update. WORK-DMD achieves superior accuracy across forecasting horizons while requiring substantially fewer total sample exposures, demonstrating the efficiency of single-pass learning.}
    \label{fig:sequential_forecasting}
\end{figure}

Table~\ref{tab:sample_efficiency} shows that WORK-DMD achieves superior accuracy across all horizons while requiring 37$\times$ fewer sample exposures than OneNet/FSNet. At short horizons ($H = 1,24$), WORK-DMD delivers approximately 2$\times$ better MSE respectively, while maintaining competitive performance. Figure~\ref{fig:sequential_forecasting} illustrates representative forecasting performance, demonstrating how WORK-DMD tracks ground truth patterns effectively with dramatically reduced sample requirements. This efficiency stems from WORK-DMD's Sherman-Morrison matrix updates that extract all necessary information in a single pass.

\begin{table}[ht]
\centering
\caption{Sample efficiency comparison in sequential online forecasting: FSNet vs. OneNet vs. WORK-DMD on Traffic dataset.}
\label{tab:sample_efficiency}
\begin{tabular}{@{} 
  c  
  c  
  c  
  c  
@{}}
\toprule
Horizon & FSNet & OneNet & WORK-DMD \\
\midrule
\multicolumn{4}{c}{\textbf{Mean Squared Error (MSE)}} \\
\midrule
$H=1$  & 0.447 & 0.855 & 0.241 \\
$H=24$ & 0.635 & 0.732 & 0.304 \\
$H=48$ & 0.597 & 0.675 & 0.448 \\
\midrule
\multicolumn{4}{c}{\textbf{Sample Efficiency}} \\
\midrule
\multicolumn{2}{c}{FSNet/OneNet Total Exposures} & \multicolumn{2}{c}{5,940} \\
\multicolumn{2}{c}{WORK-DMD Total Exposures} & \multicolumn{2}{c}{160} \\
\multicolumn{2}{c}{Efficiency Ratio} & \multicolumn{2}{c}{37.1$\times$} \\
\bottomrule
\end{tabular}
\end{table}

\section{Discussion}\label{sec:Discussion}

WORK-DMD addresses critical challenges in real-world streaming applications where traditional forecasting methods are inadequate due to computational constraints and data limitations. Our comprehensive evaluation across four benchmark datasets demonstrates that WORK-DMD consistently achieves competitive or superior performance compared to state-of-the-art online forecasting methods, with particularly strong results in short-term forecasting where it outperforms existing approaches across most datasets.

The sample efficiency analysis reveals compelling practical advantages: WORK-DMD achieves superior accuracy while requiring dramatically fewer sample exposures than competing deep learning methods. This efficiency directly translates into reduced data collection costs, faster deployment timelines, and lower computational overhead. For applications such as smart traffic management (800+ monitoring stations) or energy grid monitoring, this efficiency may be viable for real-time processing on edge hardware.

The cumulative MSE analyses provide insights into WORK-DMD's adaptability in non-stationary environments. The results demonstrate adaptation to regime shifts in both weather and electrical transformer datasets, where the method recovers from dramatic changes that would typically require manual intervention in conventional approaches. This adaptability stems from the online Sherman-Morrison updates that automatically discard outdated information while incorporating new dynamics.

From a methodological perspective, several design choices impact performance and deployment considerations. Window size selection presents a fundamental trade-off between adaptation speed and pattern retention, with smaller windows enabling rapid response to changes but potentially missing longer-term trends. Decoder update frequency represents another practical consideration: updating after every prediction ensures maximum accuracy but increases computational load, while periodic updates based on domain knowledge or systematic approaches provide more efficient alternatives. Eigenvalue monitoring offers a principled approach to system health assessment and model interpretability. Unlike black-box deep learning methods, DMD provides access to eigenvalue analysis, which encodes information about system dynamics, dominant frequencies, and temporal patterns \citep{kutz2016dmd}. This spectral decomposition enables systematic approaches to detecting when model retraining or parameter adjustment is necessary. Recent work has demonstrated that DMD-based methods can be leveraged for change point detection by monitoring eigenvalue drift and reconstruction errors \citep{khamesi2024online}, suggesting potential extensions of WORK-DMD for automated system monitoring. However, developing robust online change point detection methods from eigenvalue patterns remains an active area of research.

The computational advantages extend beyond efficiency metrics to enable new deployment paradigms. WORK-DMD's Sherman-Morrison updates operate with lower computational complexity per timestep compared to full matrix operations, while maintaining constant memory requirements rather than growing with data stream length. These properties could enable deployment on embedded systems and IoT devices, where traditional neural networks might exceed hardware constraints.

However, some limitations constrain the applicability of the method. As a system identification approach, WORK-DMD assumes underlying dynamical structure that may not exist in purely stochastic processes such as random walks, white noise processes, or series dominated by measurement noise without underlying dynamics \citep{brunton2022data}. The method excels with structured time series that exhibit recognizable patterns, such as traffic flows, energy consumption, and sensor readings, but may struggle with highly irregular or chaotic dynamics. Additionally, hyperparameter optimization remains a practical challenge, particularly for automated deployment scenarios where domain expertise is limited.

Future research should address practical deployment challenges for widespread industrial adoption. Adaptive hyperparameter tuning mechanisms that respond to changing data characteristics could enhance robustness across diverse applications. Integration with anomaly detection systems would enable automatic model reset when the underlying dynamics shift dramatically. Extension to explicit multivariate forecasting with cross-variable modeling could expand applicability to complex systems such as chemical process control, smart manufacturing, and energy management where variable interactions drive system behavior \citep{tang2023gru, sitapure2024time, xiang2025flexible}.

The advantages of computational efficiency support sustainable AI deployment strategies by reducing resource requirements compared to traditional deep learning approaches. The method's ability to operate with limited data collection addresses privacy concerns in sensitive applications while enabling predictive capabilities in data-constrained environments. However, the interpretability advantages of eigenvalue evolution must be balanced against risks of over-reliance on automated decisions in critical infrastructure, emphasizing the importance of human oversight protocols in safety-critical deployments.

\section{Conclusion}
\label{sec:Conclusion}

We present WORK-DMD, a method that integrates Random Fourier Features with online Dynamic Mode Decomposition to enable real-time forecasting from streaming data. By capturing nonlinear dynamics through explicit feature mapping while maintaining fixed computational cost per update, WORK-DMD transforms traditional kernel methods into a practical online learning framework. 

Experimental evaluation shows WORK-DMD achieves competitive or superior accuracy compared to recent online forecasting methods across benchmark datasets while requiring only a single pass through the data. The method excels in short-term forecasting scenarios where rapid adaptation is critical. Unlike transformer-based models that require substantial computational resources and lengthy training, WORK-DMD generates reliable predictions from the current windowed snapshot alone. This efficiency could make it suitable for resource-constrained environments such as edge devices and IoT sensors.

By augmenting dynamical systems theory with kernel approximation techniques, WORK-DMD demonstrates that sophisticated nonlinear modeling need not require deep learning's computational overhead. The combination of efficiency, minimal data requirements, and interpretable eigenvalue diagnostics positions WORK-DMD as a valuable tool for applications demanding real-time adaptation under strict computational constraints.


\section*{Appendix A}
\label{sec:sensitivity}

To evaluate the hyperparameter robustness of WORK-DMD, we conducted a systematic sensitivity analysis of four critical hyperparameters: window size, Random Fourier Feature (RFF) dimension, kernel bandwidth parameter ($\gamma$), and matrix rank. We varied each parameter independently while maintaining baseline values (window size = 60, RFF dimension = 1,024, $\gamma = 3 \times 10^{-5}$, rank = 128) for others. Each configuration was evaluated using training data of the forecasts on the Traffic dataset to ensure statistical reliability.

\begin{figure}[htbp]
    \centering
    \begin{subfigure}[b]{\textwidth}
        \includegraphics[width=\textwidth]{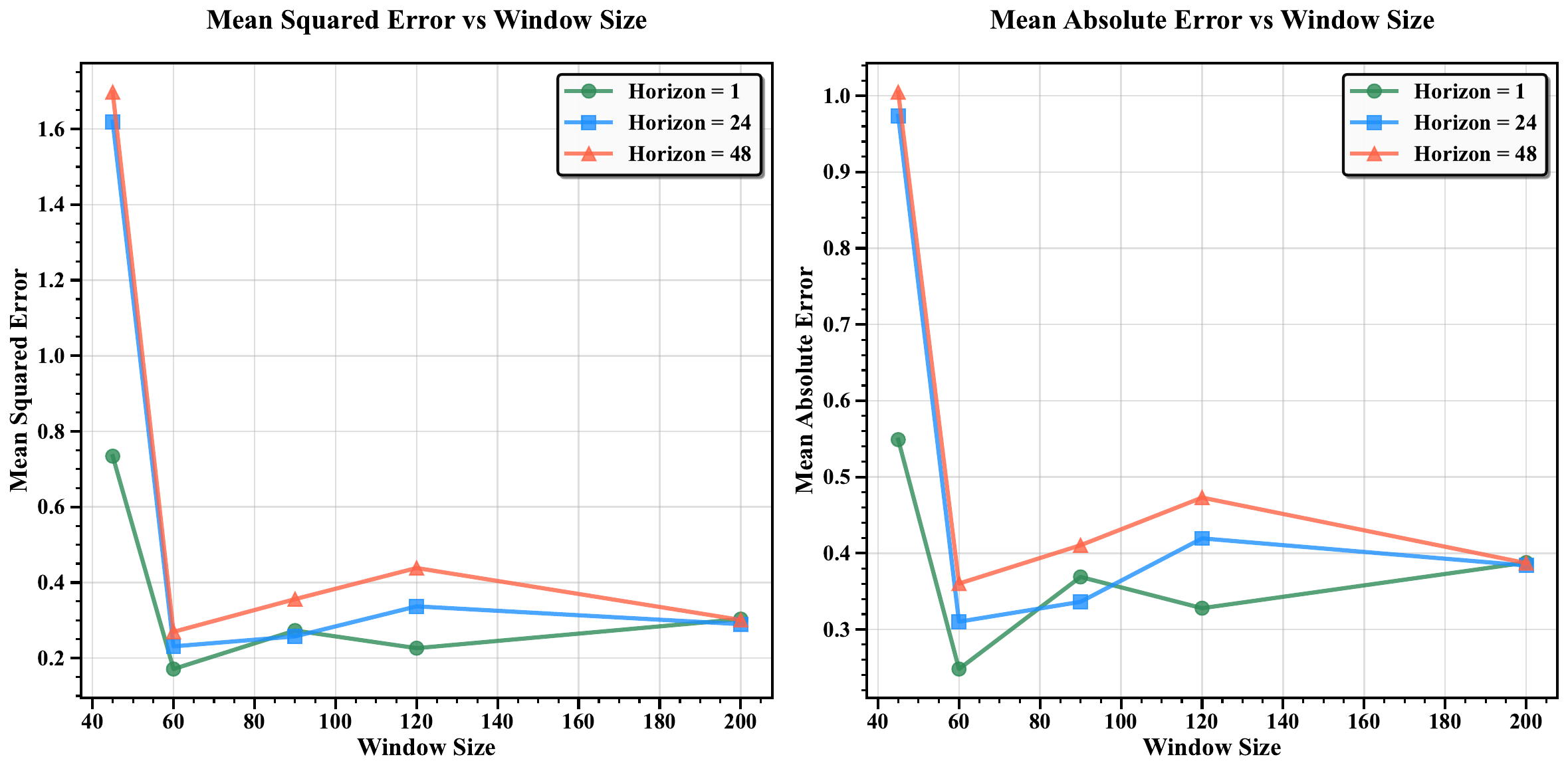}
        \caption{Window Size Sensitivity}
        \label{fig:sens_window}
    \end{subfigure}
    
    \vspace{0.8cm}
    
    \begin{subfigure}[b]{\textwidth}
        \includegraphics[width=\textwidth]{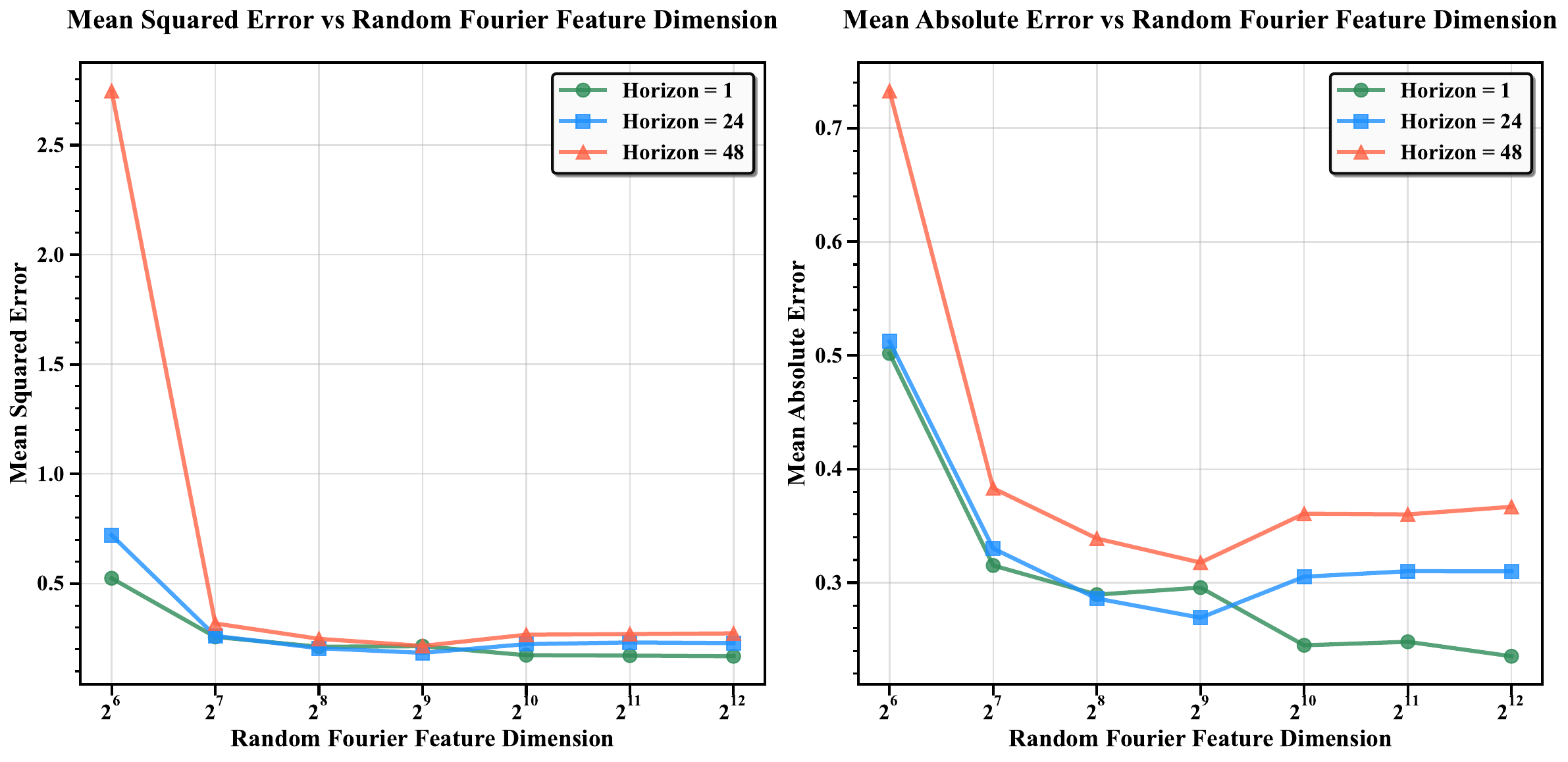}
        \caption{RFF Dimension Sensitivity}
        \label{fig:sens_rff}
    \end{subfigure}
    
    \caption{Window size and RFF dimension sensitivity analysis for WORK-DMD showing MSE and MAE performance across different forecasting horizons ($H$=1, 24, 48). (a) Window size analysis reveals optimal performance around 60-90 time steps, balancing temporal context with computational efficiency. Smaller windows provide insufficient historical information while larger windows introduce computational overhead without proportional gains. (b) RFF dimension shows diminishing returns beyond 1024 features, with plateau behavior indicating sufficient kernel approximation quality.}
    \label{fig:sensitivity_analysis_1}
\end{figure}

\begin{figure}[htbp]
    \centering
    \begin{subfigure}[b]{\textwidth}
        \includegraphics[width=\textwidth]{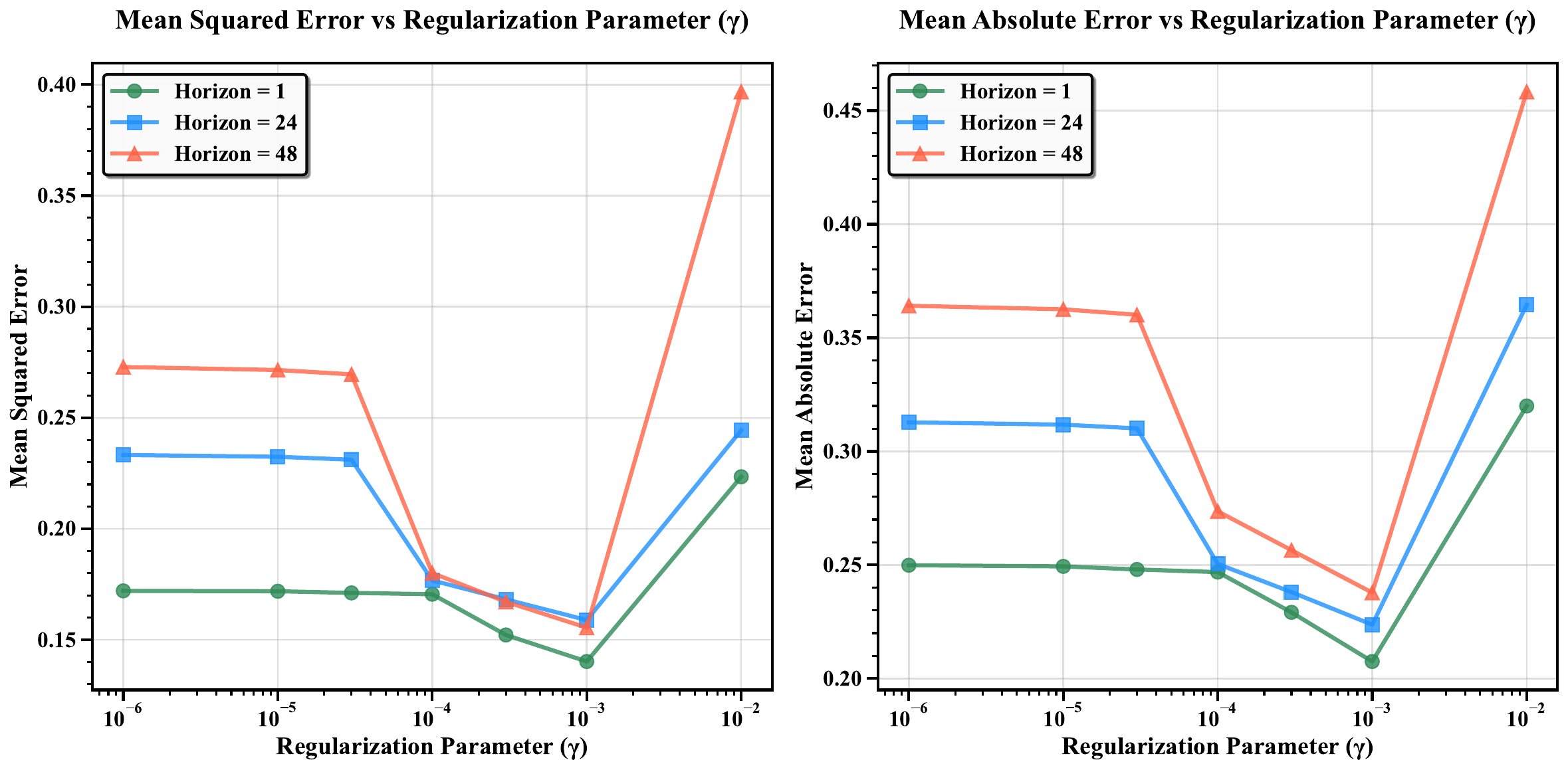}
        \caption{Gamma Parameter Sensitivity}
        \label{fig:sens_gamma}
    \end{subfigure}
    
    \vspace{0.8cm}
    
    \begin{subfigure}[b]{\textwidth}
        \includegraphics[width=\textwidth]{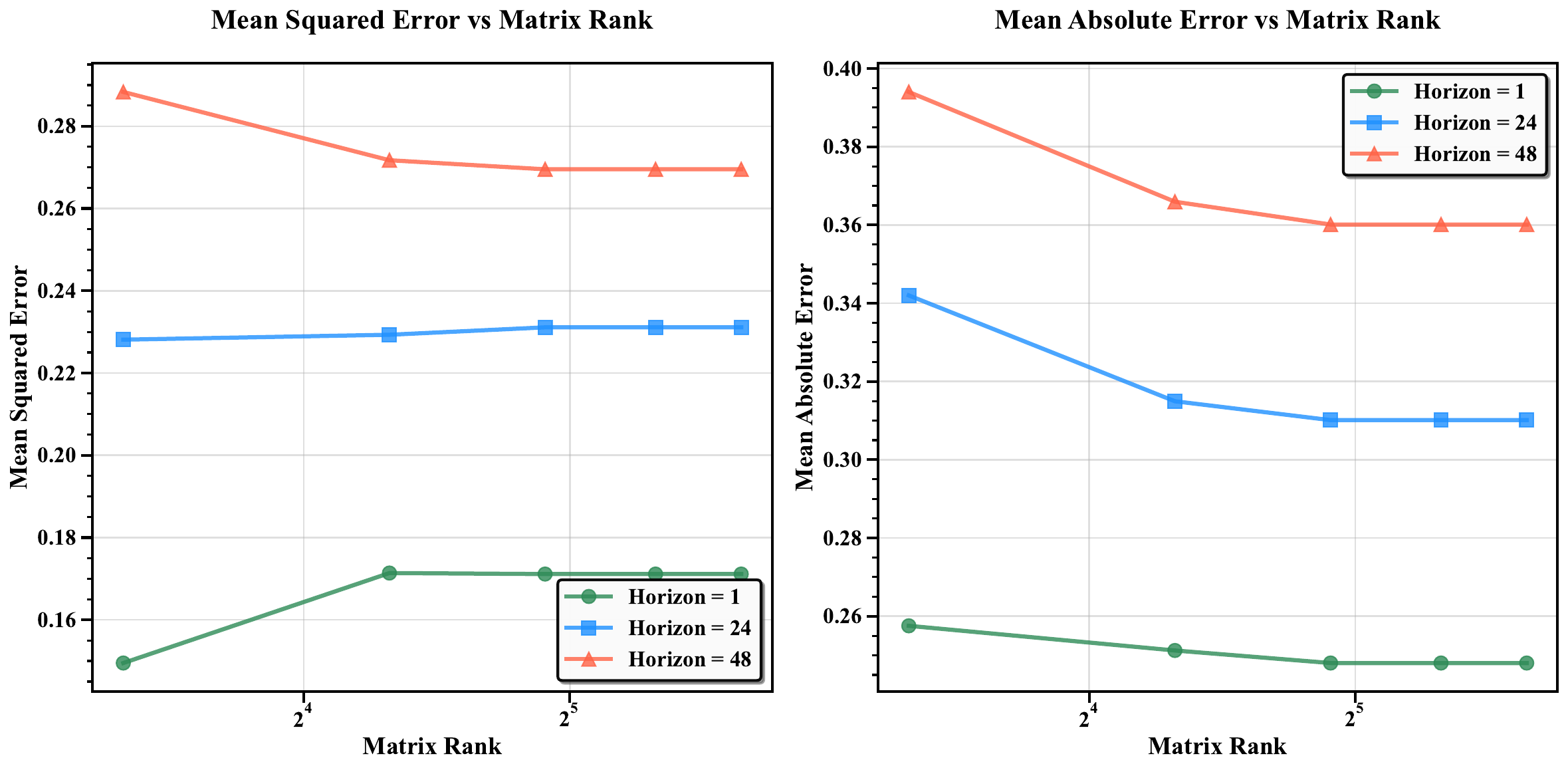}
        \caption{Rank Sensitivity}
        \label{fig:sens_rank}
    \end{subfigure}
    
    \caption{Gamma parameter and rank sensitivity analysis for WORK-DMD. (c) Gamma parameter exhibits optimal range around $10^{-5}$ to $10^{-4}$, with performance degradation at extreme values due to under/over-smoothing effects. This parameter shows the highest sensitivity among all tested hyperparameters. (d) Rank truncation demonstrates stable performance between 20-40, with degradation below 20 due to insufficient model complexity and above 40 due to overfitting in the streaming setting.}
    \label{fig:sensitivity_analysis_2}
\end{figure}

The analysis reveals distinct sensitivity patterns across parameters. Window size exhibits optimal performance around 60-90 time steps, where smaller windows provide insufficient temporal context while larger windows introduce computational overhead without accuracy gains (Figure~\ref{fig:sensitivity_analysis_1}(a)). RFF dimension demonstrates diminishing returns beyond 1024 features, with performance plateauing due to sufficient kernel approximation quality (Figure~\ref{fig:sensitivity_analysis_1}(b)). The kernel bandwidth parameter $\gamma$ shows the highest sensitivity, with optimal values around $10^{-5}$ to $10^{-4}$; values outside this range cause either over-smoothing or noise amplification (Figure~\ref{fig:sensitivity_analysis_2}(a)). Matrix rank exhibits moderate sensitivity with stable performance between 20-40, where lower ranks lack model expressiveness and higher ranks introduce overfitting in the streaming setting (Figure~\ref{fig:sensitivity_analysis_2}(b)).

These results demonstrate that WORK-DMD exhibits reasonable robustness within practical hyperparameter ranges, with the kernel bandwidth requiring the most careful tuning. The identified optimal ranges facilitate deployment across different domains: window size should match the temporal correlation structure of the application, RFF dimensions between 512-2048 provide good accuracy-efficiency balance, $\gamma$ should be tuned within $[10^{-6}, 10^{-4}]$ using validation data, and rank values between 20-40 offer sufficient model complexity without overfitting.

\section*{Appendix B}
\label{sec: Traffic Forecast}

\begin{figure*}[htbp]
    \centering
    \begin{subfigure}[b]{0.48\textwidth}
        \centering
        \includegraphics[width=\textwidth]{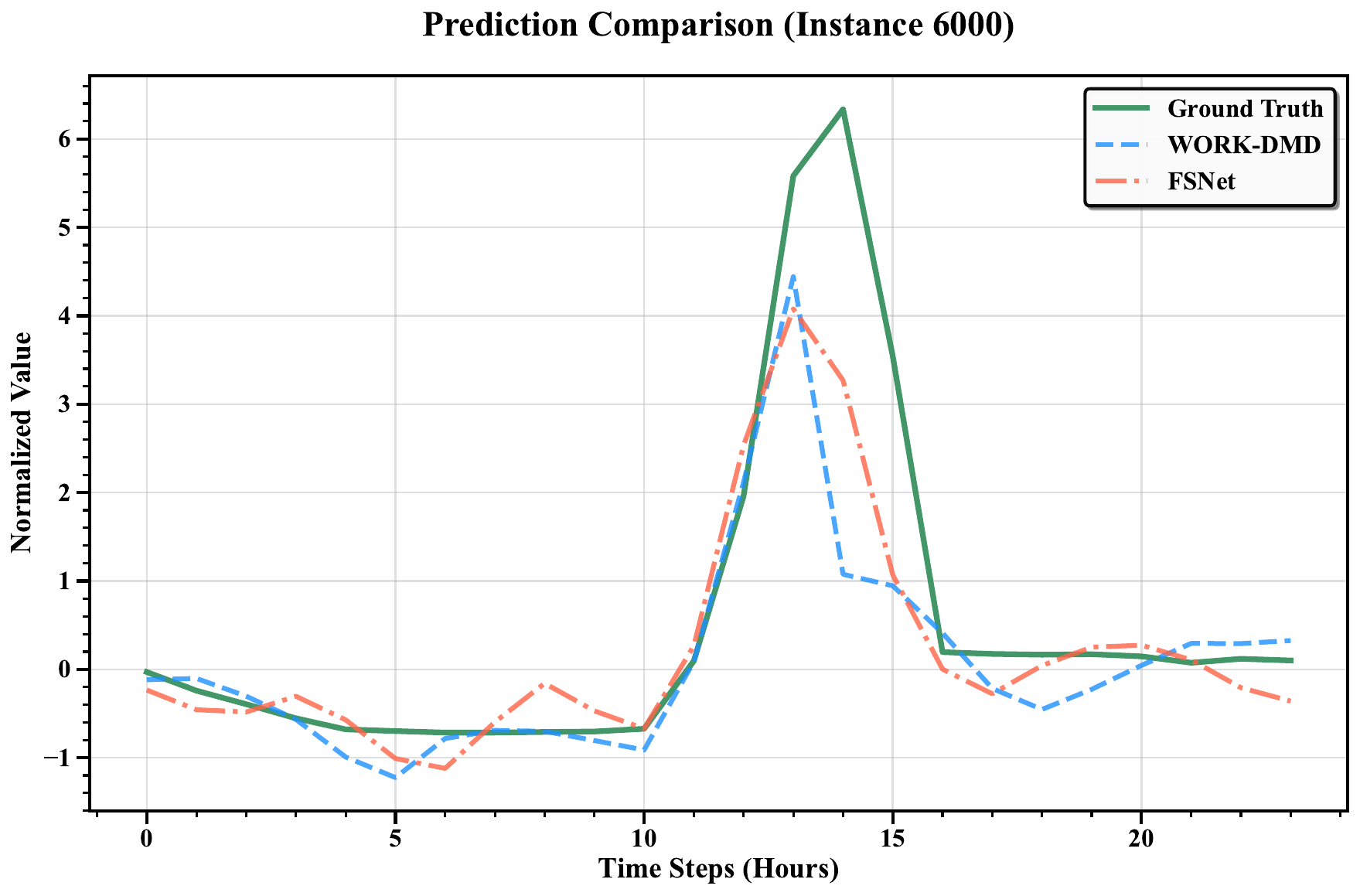}
        \caption{Channel 0 - Instance 6000}
        \label{fig:traffic_comparison_6000_ch0}
    \end{subfigure}
    \hfill
    \begin{subfigure}[b]{0.48\textwidth}
        \centering
        \includegraphics[width=\textwidth]{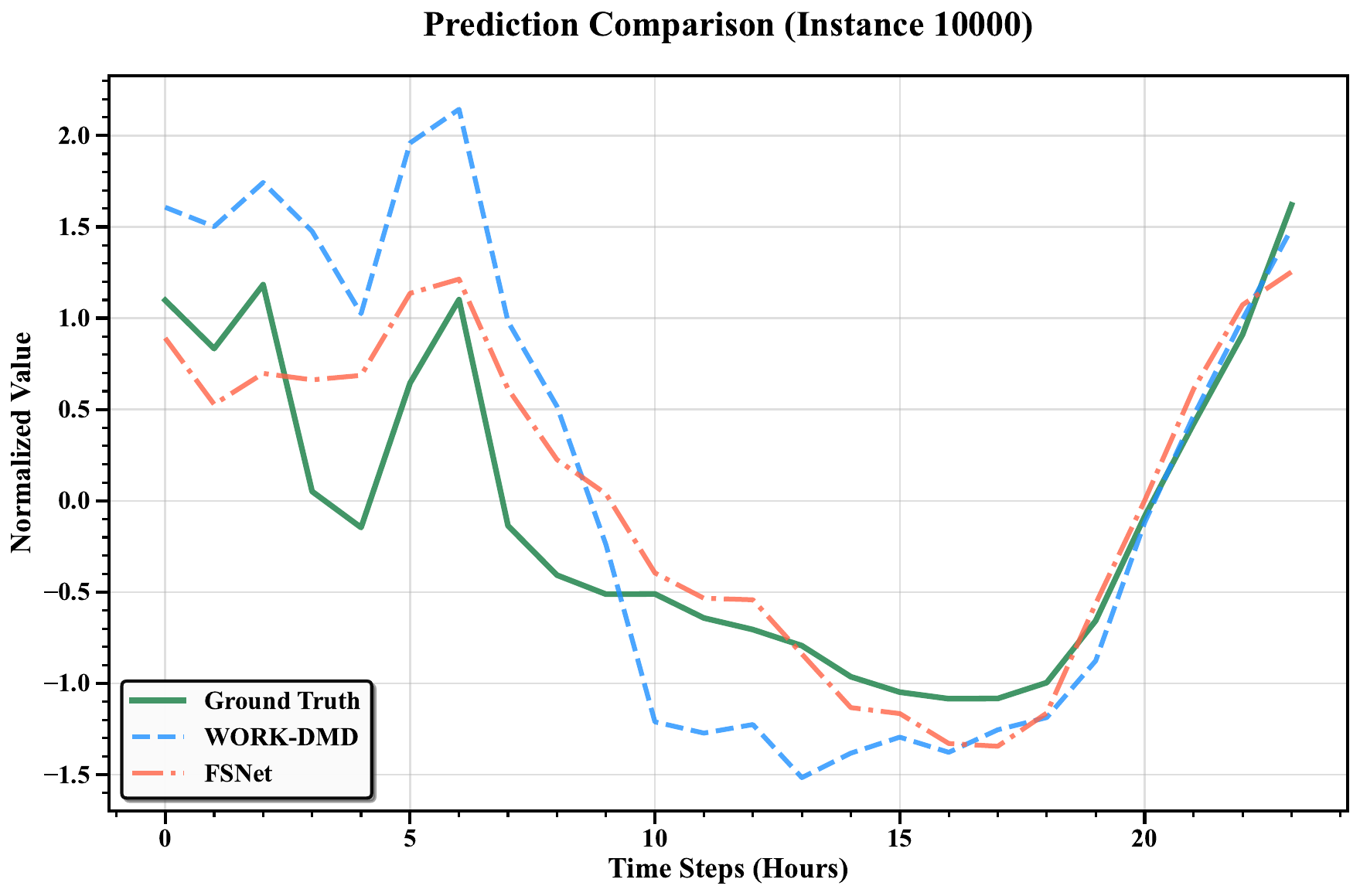}
        \caption{Channel 0 - Instance 10000}
        \label{fig:traffic_comparison_10000_ch0}
    \end{subfigure}
    
    \vspace{0.3cm}
    
    \begin{subfigure}[b]{0.48\textwidth}
        \centering
        \includegraphics[width=\textwidth]{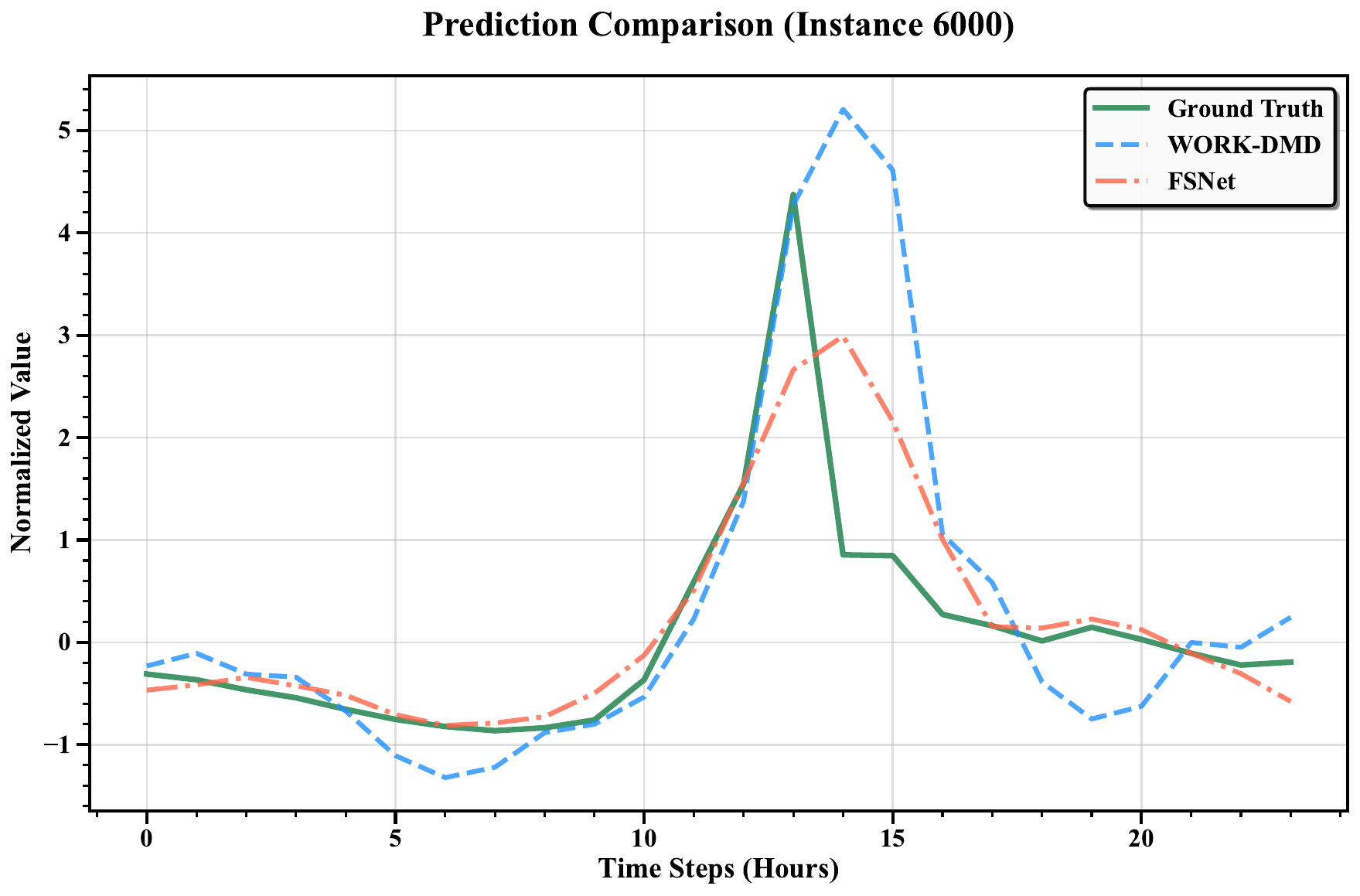}
        \caption{Channel 1 - Instance 6000}
        \label{fig:traffic_comparison_6000_ch1}
    \end{subfigure}
    \hfill
    \begin{subfigure}[b]{0.48\textwidth}
        \centering
        \includegraphics[width=\textwidth]{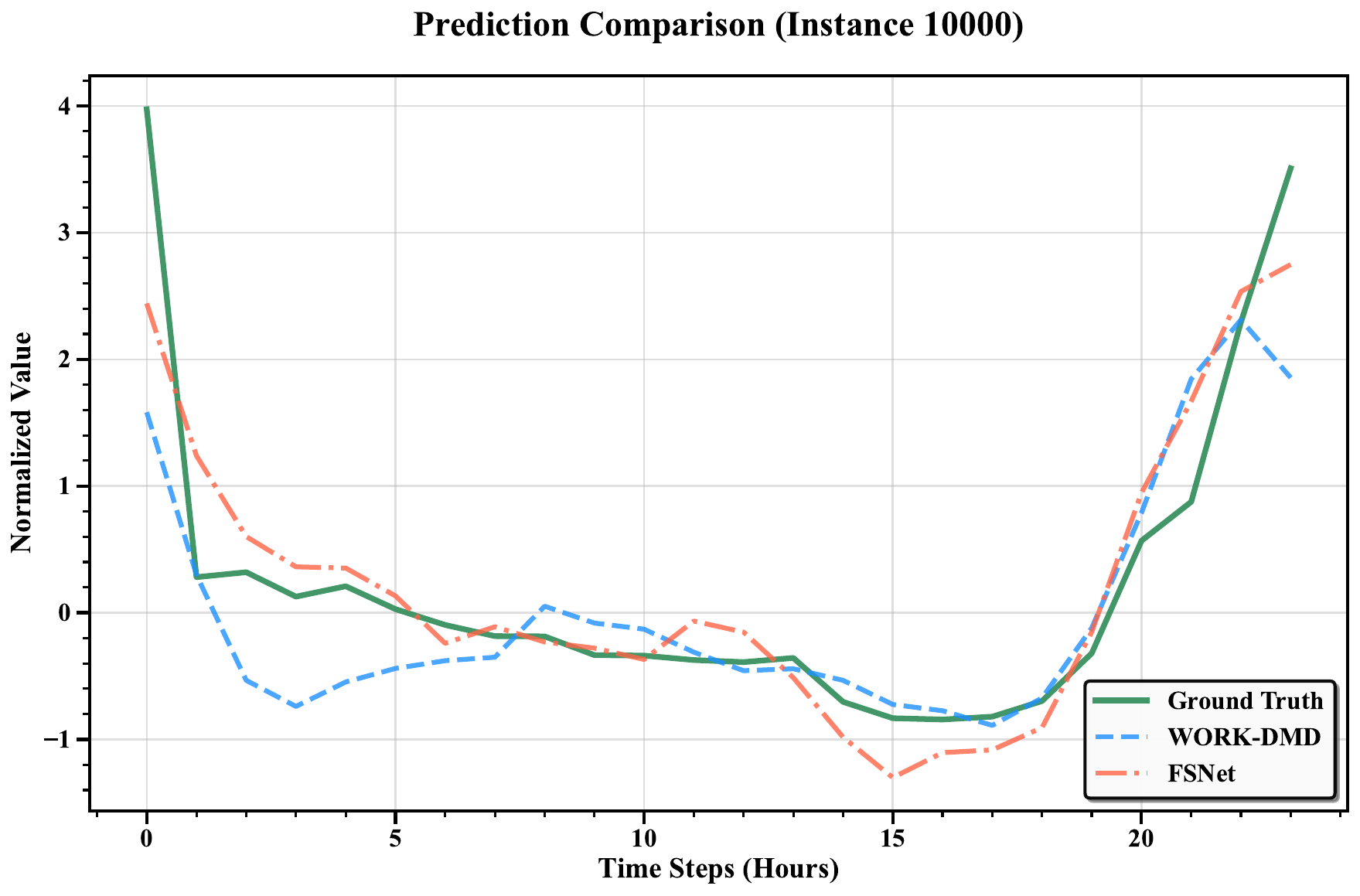}
        \caption{Channel 1 - Instance 10000}
        \label{fig:traffic_comparison_10000_ch1}
    \end{subfigure}
    
    \vspace{0.3cm}
    
    \begin{subfigure}[b]{0.48\textwidth}
        \centering
        \includegraphics[width=\textwidth]{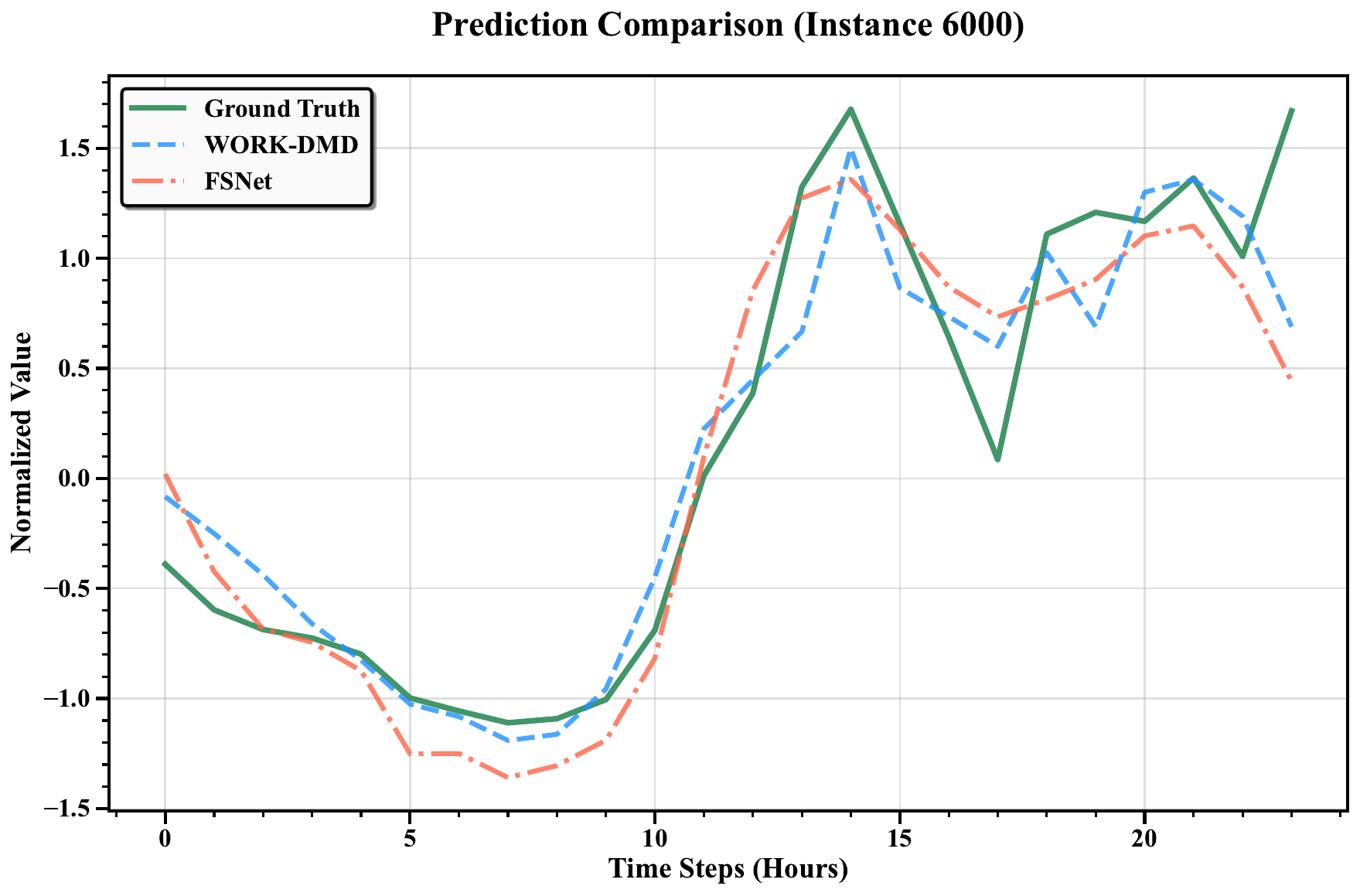}
        \caption{Channel 2 - Instance 6000}
        \label{fig:traffic_comparison_6000_ch2}
    \end{subfigure}
    \hfill
    \begin{subfigure}[b]{0.48\textwidth}
        \centering
        \includegraphics[width=\textwidth]{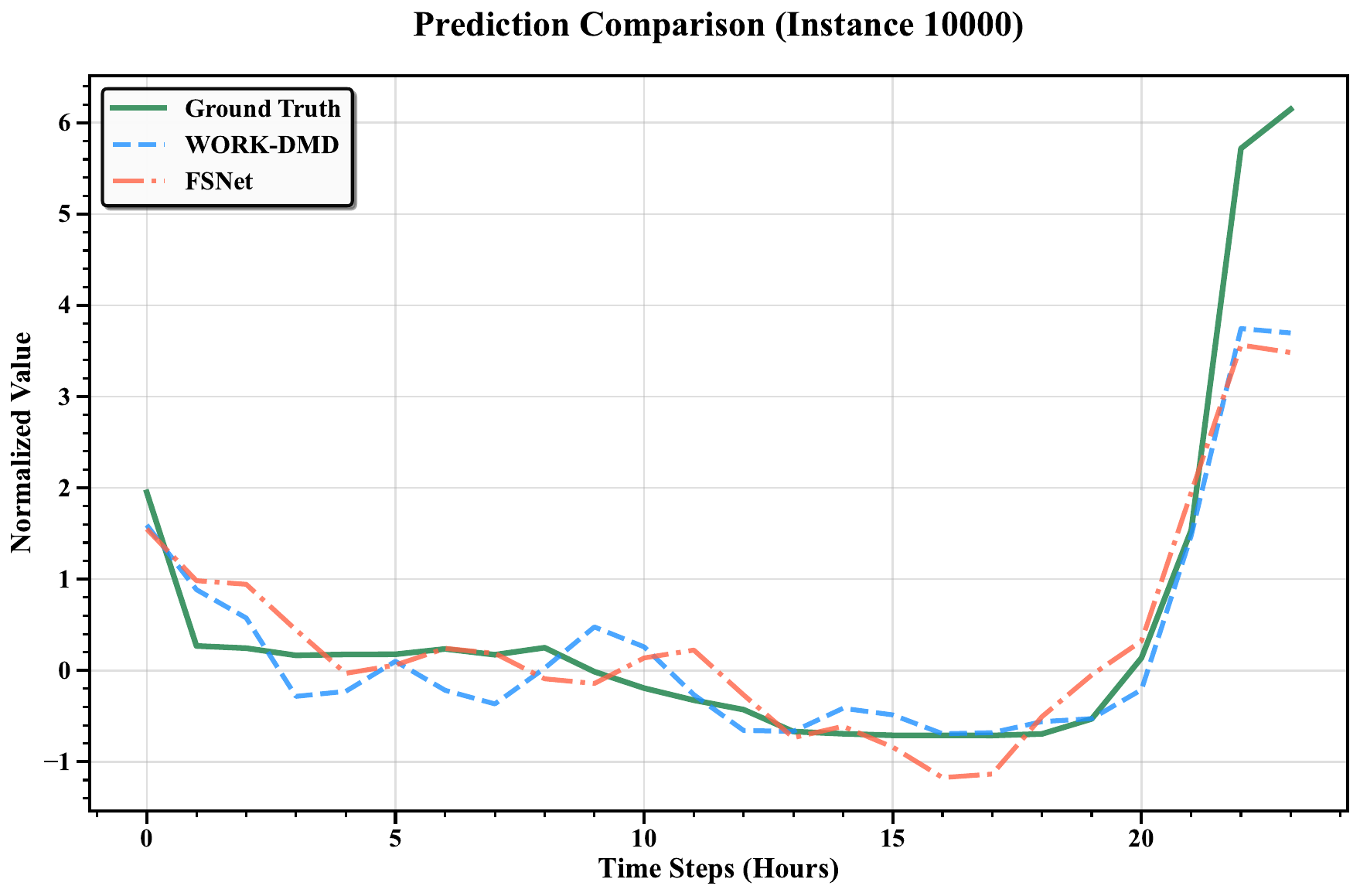}
        \caption{Channel 2 - Instance 10000}
        \label{fig:traffic_comparison_10000_ch2}
    \end{subfigure}
    
    \caption{Time series forecasting comparison on Traffic dataset for three monitoring channels at two temporal instances. The figure shows 24-step ahead predictions ($H=24$) comparing WORK-DMD and FSNet against ground truth. WORK-DMD achieves comparable performance to FSNet across all channels (only first three channels shown for clarity), demonstrating effective tracking of complex traffic dynamics.}
    \label{fig:traffic_forecasting_comparison}
\end{figure*}

When comparing forecasting plots against the latest methods, we found that OneNet did not provide results for the traffic dataset. We therefore use FSNet as a comparison baseline. Figure \ref{fig:traffic_forecasting_comparison} shows the results for the first three channels of traffic monitoring. For prediction horizon $H = 24$, both models demonstrate comparable performance in capturing the ground truth patterns.

\section*{Appendix C}
\label{sec: app_error_plots}

In addition to the cumulative error plots that were shown in section \ref{sec:Results} for the ETTh2 and WTH forecasting results, we present the cumulative results for the ETTm1 and Traffic datasets here. It should be noted, since there were no available results for OneNet with the Traffic dataset, FSNet was the next closest comparison.

The ETTm1 results demonstrate WORK-DMD's exceptional performance for short-term forecasting, with notably superior accuracy in the 1-step ahead prediction task where the cumulative error remains consistently lower than OneNet throughout the evaluation period. For longer horizons (48-step ahead), WORK-DMD maintains competitive performance with OneNet, showing stable error accumulation patterns that indicate robust forecasting capability across different prediction timeframes. The Traffic dataset comparison with FSNet reveals consistent relative stability between the two methods, with both algorithms demonstrating similar error accumulation rates across the 1-step and 24-step ahead forecasting tasks, suggesting that WORK-DMD achieves comparable performance to established baselines even on challenging high-dimensional traffic prediction problems.

\begin{figure}[htbp]
    \centering
    \begin{subfigure}[b]{0.8\textwidth}
        \centering
        \includegraphics[width=\textwidth]{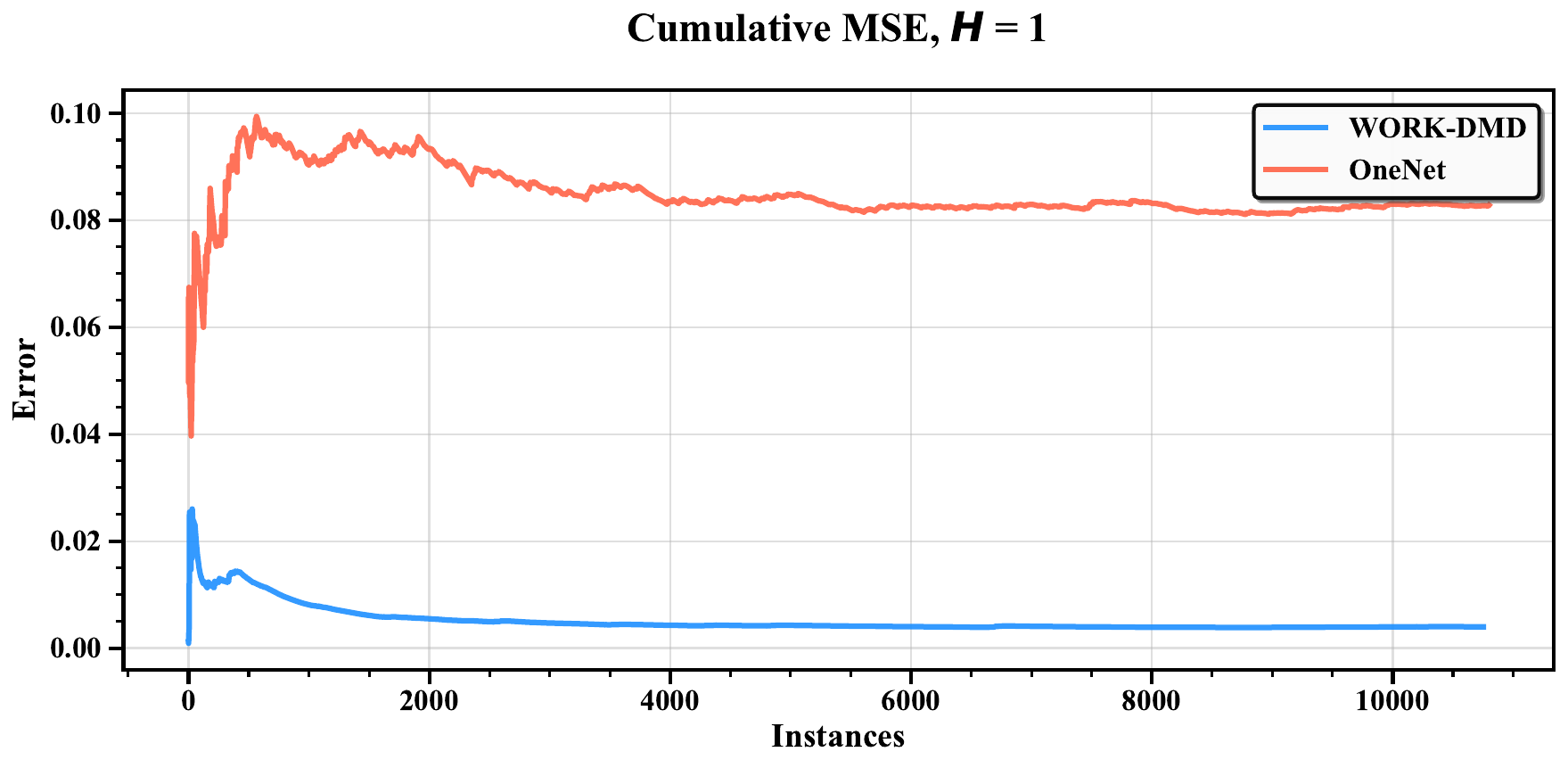}
        \caption{1-step ahead}
        \label{fig:ETTm1_cum_1}
    \end{subfigure}
    
    \vspace{0.4cm}
    
    \begin{subfigure}[b]{0.8\textwidth}
        \centering
        \includegraphics[width=\textwidth]{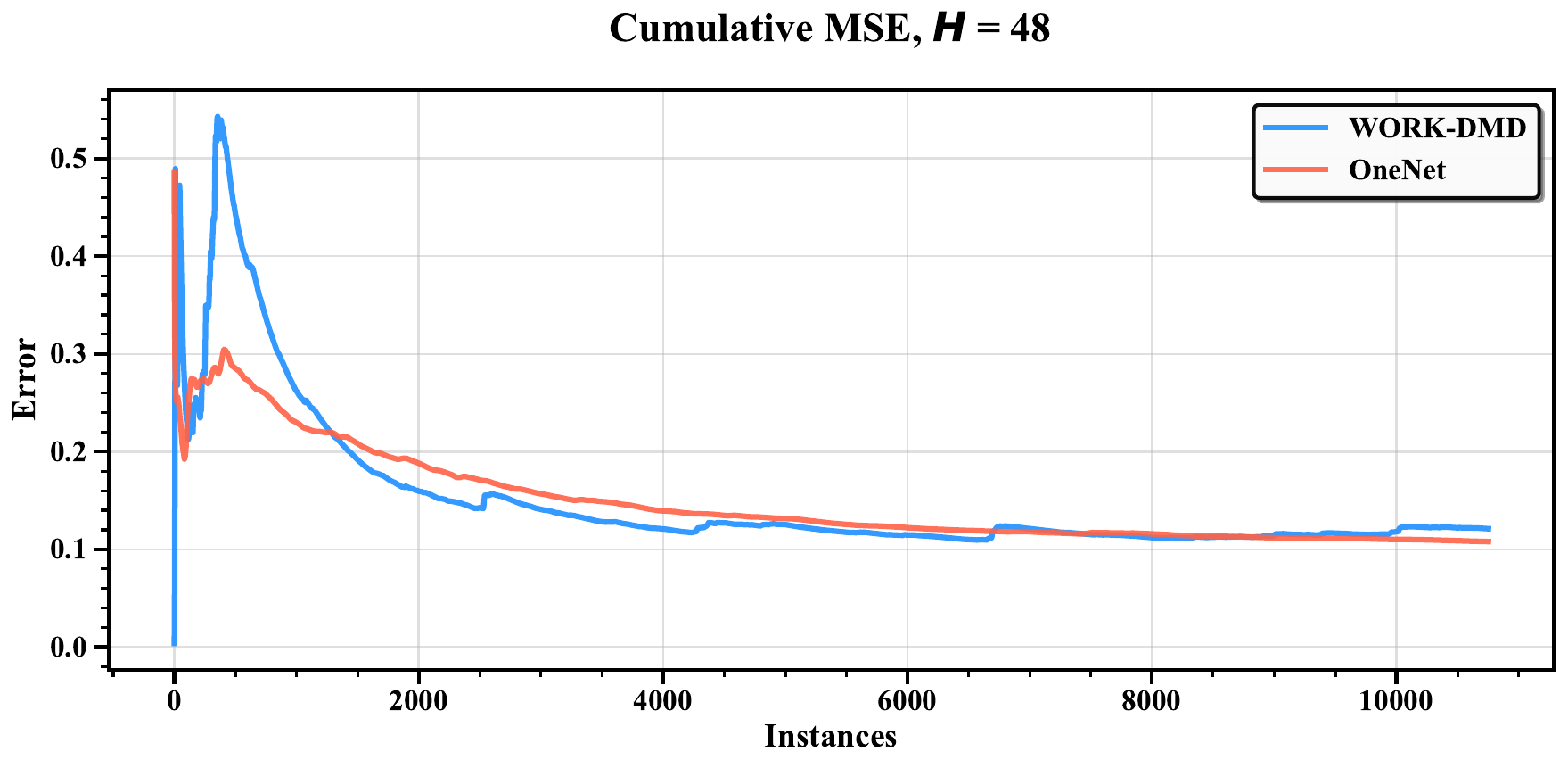}
        \caption{48-step ahead}
        \label{fig:ETTm1_cum_48}
    \end{subfigure}
    
    \caption{Cumulative MSE comparison on ETTm1 dataset. The figures show cumulative mean squared error progression for 1-step and 48-step ahead forecasting, comparing WORK-DMD and OneNet performance over time. The cumulative curves demonstrate superior performance when $H = 1$ and competitive performance when $H = 48$.}
    \label{fig:ETTm1_cumulative_MSE}
\end{figure}

\begin{figure}[htbp]
    \centering
    \begin{subfigure}[b]{0.8\textwidth}
        \centering
        \includegraphics[width=\textwidth]{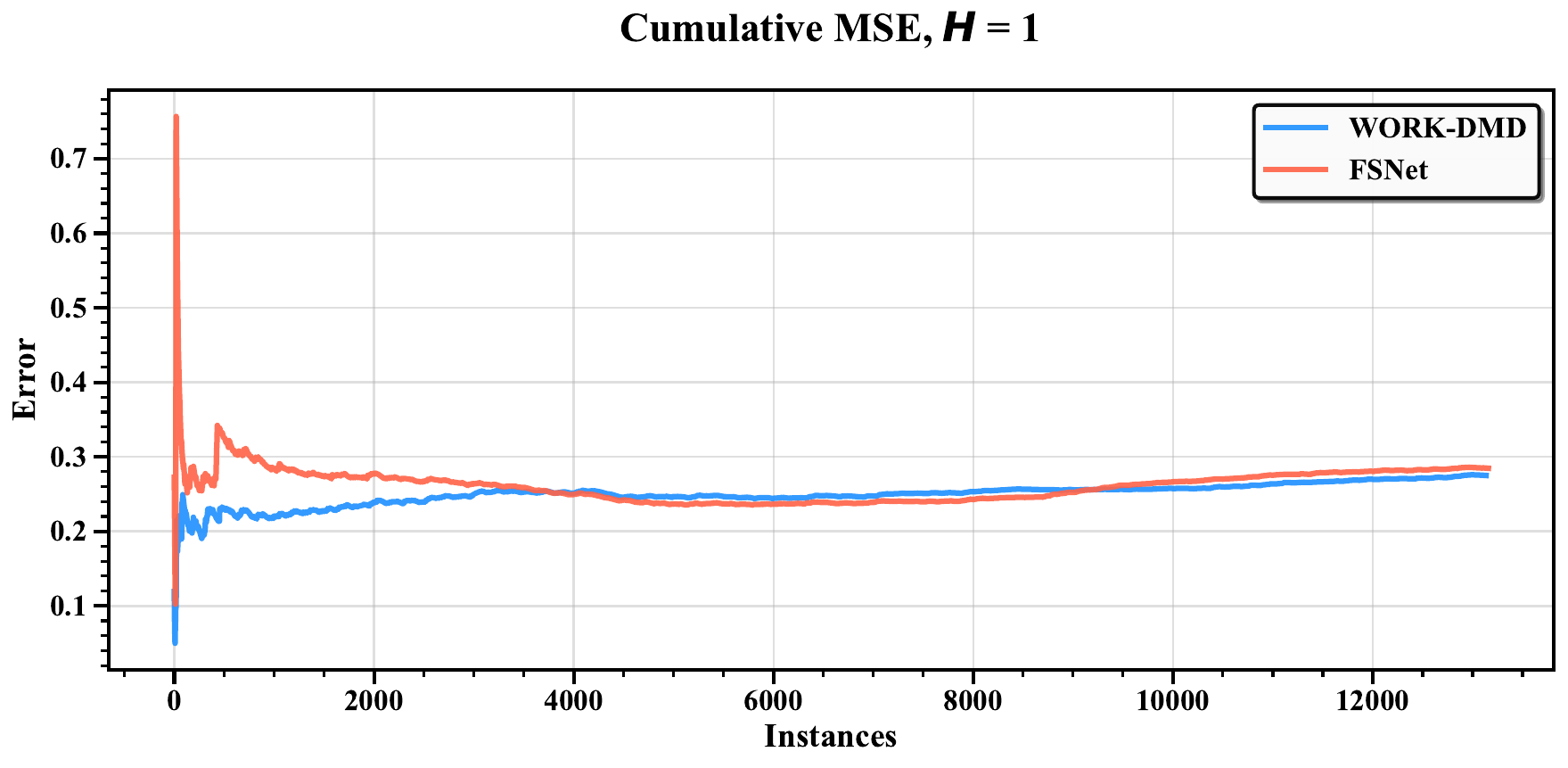}
        \caption{1-step ahead}
        \label{fig:Traffic_cum_1}
    \end{subfigure}
    
    \vspace{0.4cm}
    
    \begin{subfigure}[b]{0.8\textwidth}
        \centering
        \includegraphics[width=\textwidth]{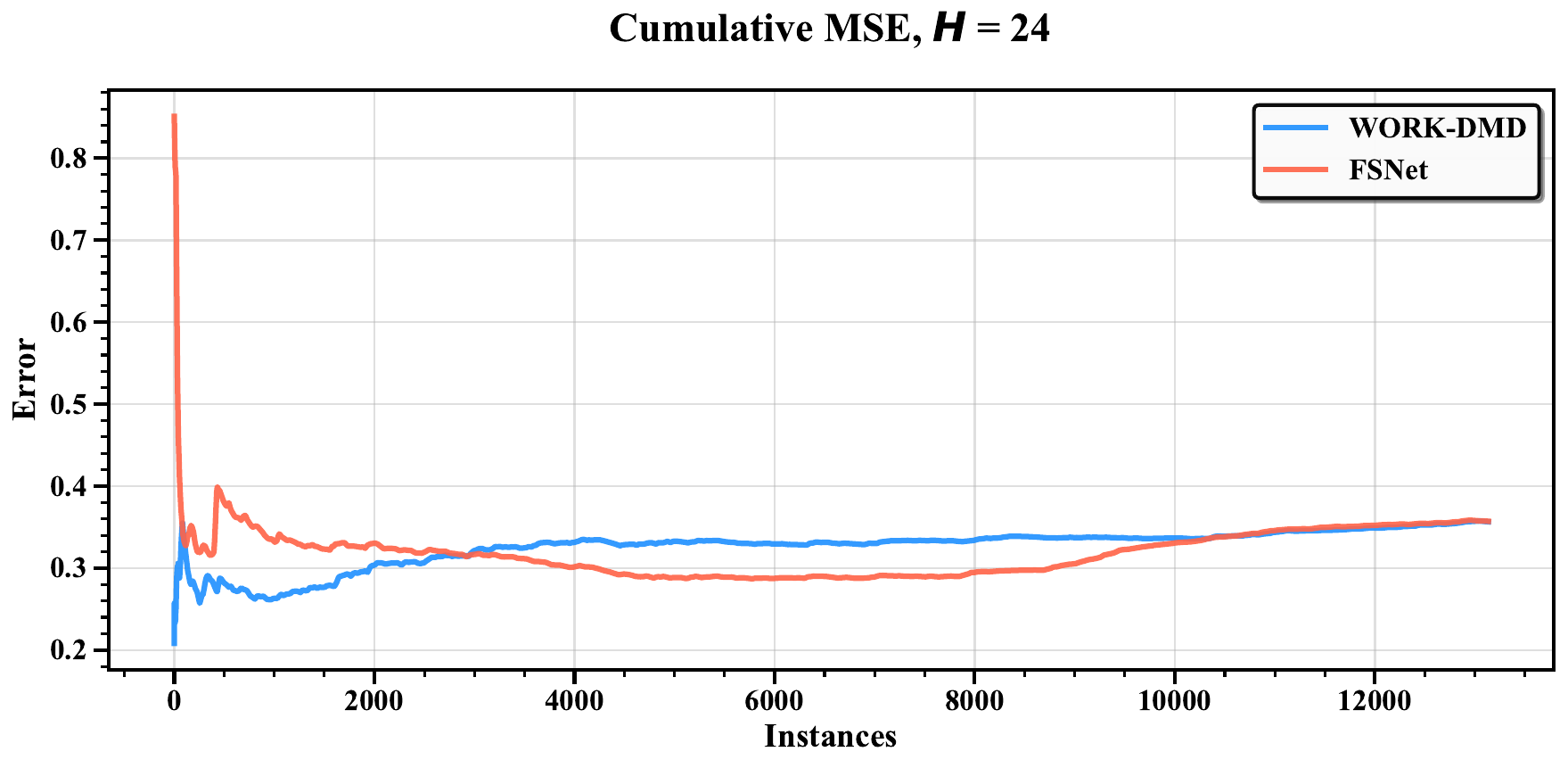}
        \caption{24-step ahead}
        \label{fig:Traffic_cum_24}
    \end{subfigure}
    
    \caption{Cumulative MSE comparison on Traffic dataset. The figures show cumulative mean squared error progression for ahead forecasting, comparing WORK-DMD and FSNet performance. The cumulative analysis reveals the relative stability and accuracy of both methods across different prediction horizons.}
    \label{fig:Traffic_cumulative_MSE}
\end{figure}




\bibliographystyle{informs2014} 
\bibliography{references} 

\end{document}